\documentclass[11pt]{article}
\usepackage{tcolorbox}

\usepackage{textcomp}
\usepackage{newunicodechar}
\newunicodechar{≈}{\ensuremath{\approx}}
\usepackage[utf8]{inputenc}
\usepackage{textcomp}
\usepackage{textgreek}

\usepackage[preprint]{acl}

\usepackage{times}
\usepackage{latexsym}
\usepackage{graphicx}
\usepackage{amsmath}
\usepackage{amssymb}
\usepackage{booktabs}
\usepackage{wrapfig} 
\usepackage{xcolor}
\usepackage{tcolorbox}
\tcbuselibrary{skins,breakable}
\usepackage{enumitem}
\usepackage{listings}
\usepackage{tabularx}
\usepackage{array}
\usepackage{ragged2e}
\usepackage{multirow}
\usepackage{caption}
\usepackage[export]{adjustbox}
\usepackage{subcaption}
\usepackage{multirow}
\usepackage{xurl}  

\newcolumntype{L}[1]{>{\RaggedRight\arraybackslash}p{#1}}
\newcolumntype{Y}{>{\RaggedRight\arraybackslash}X}
\newcolumntype{R}[1]{>{\RaggedLeft\arraybackslash}p{#1}}

\renewcommand{\arraystretch}{1.18}
\makeatletter
\newcommand{\sidecaption}[2]{%
    \refstepcounter{figure}%
    \@makecaption{\figurename~\thefigure}{#1}%
    \label{#2}%
}
\makeatother

\usepackage[T1]{fontenc}

\usepackage[utf8]{inputenc}

\usepackage{microtype}

\usepackage{inconsolata}

\usepackage{graphicx}
\usepackage{hyperref}

\title{\texttt{DRIP-R}: A Benchmark for Decision-Making and Reasoning Under Real-World Policy Ambiguity in the Retail Domain}

\author{
  Hsuvas Borkakoty\thanks{Equal contribution.}$^{1}$
  \quad
  Sebastian Pohl\footnotemark[1]$^{1}$
  \quad
  Cheng Wang$^{2}$
  \quad
  Bei Chen$^{2}$
  \quad
  Yufang Hou$^{1}$
  \\[4pt]
  $^{1}$IT:U Interdisciplinary Transformation University Austria
  \\
  $^{2}$Amazon, Berlin, Germany
  \\[4pt]
  \texttt{\{hsuvas.borkakoty, sebastian.pohl, yufang.hou\}@it-u.at}
  \\
  \texttt{\{cwngam, chenbe\}@amazon.com}
}

\begin{document}
\maketitle
\begin{abstract}
LLM-based agents are increasingly deployed for routine but consequential tasks in real-world domains, where their behavior is governed by inherently ambiguous domain policies that admit multiple valid interpretations. Despite the prevalence of such ambiguities in practice, existing agent benchmarks largely assume unambiguous, well-specified policies, leaving a critical evaluation gap. We introduce \texttt{DRIP-R}, a benchmark that systematically exploits real-world retail policy ambiguities to construct scenarios in which no single correct resolution exists. \texttt{DRIP-R} comprises a curated set of policy-ambiguous return scenarios paired with a realistic customer personas, a full-duplex conversational simulation with tool-calling capabilities and a multi-judge evaluation framework covering policy adherence, dialogue quality, behavioral alignment, and resolution quality. Our experiments show that frontier models fundamentally disagree on identical policy-ambiguous scenarios, confirming that ambiguity poses a genuine and systematic challenge to LLM decision-making.\footnote{The benchmark can be accessed at \href{https://github.com/ITU-NLP/drip-r-bench}{This URL}} 
\end{abstract}
\section{Introduction}
\label{sec:intro}

LLM-based conversational agents are increasingly deployed in real-world domains such as retail, healthcare, and finance to handle high-stakes, domain-specific tasks \citep{luo2025large}. In these settings, domain policies play a central role: they specify which actions are permissible, constrain expected behavior across task scenarios, and provide the normative basis for judging whether an agent’s output constitutes an appropriate resolution. 

In parallel, recent advances in LLM agent benchmarking have improved our understanding of agent capabilities across multiple dimensions \citep{yehudai2025survey}. 
Existing agent evaluation work primarily focuses on agents' abilities 
required to resolve tasks, including planning \citep{zhang2024probing}, instruction following \citep{wang2026trajectory2task, sun2026ambibench}, and tool-calling \citep{lu2025toolsandbox,chen2025acebench}. 
However, the policies used in these benchmarks are often purpose-built for evaluation. They are cleanly specified, narrowly scoped, and tailored to the benchmark tasks. Although this design makes evaluation tractable, it abstracts away a central difficulty of real-world deployment, i.e., real policies are rarely complete and unambiguous.

\begin{figure*}[t]
    \centering
    \includegraphics[width=\textwidth]{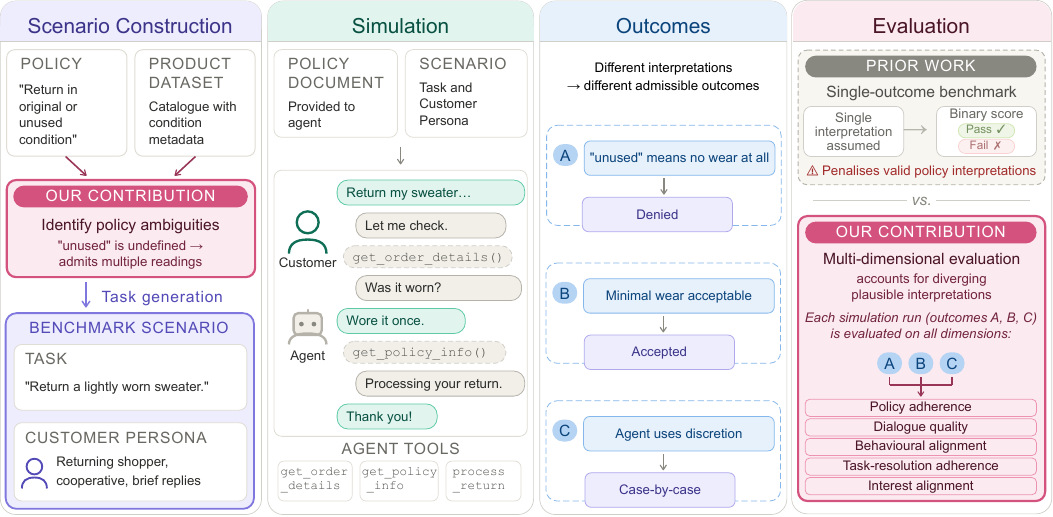}
    \caption{\textbf{Benchmark overview.} The scenarios are built from policy ambiguities: return tasks resolvable in multiple ways depending on policy interpretation and resolved through conversation between customer and service agent. } 

    \label{fig:overall_benchmark}
\end{figure*}


Real-world domain policies, 
often contain  implicit assumptions and ambiguous language  
that admit multiple valid interpretations \citep{fowler2023strategies}. For example, the statement \textit{`Items can be returned as long as they are in unused condition'} from Amazon's return policy\footnote{\href{https://www.amazon.com/gp/help/customer/display.html?nodeId=GKM69DUUYKQWKWX7}{Amazon order return policy}} opens the question of what constitutes \textit{`unused'} for a specific item, leading to different yet defensible conclusions about an item's return eligibility. These ambiguities can be classified as \emph{vagueness}, \emph{semantic ambiguity}, \emph{referential ambiguity}, or \emph{incompleteness} \citep{massey2014identifying}.  


These ambiguities reflect a deeper real-world challenge. When legitimate values conflict (such as customer satisfaction, organizational profit, and regulatory compliance), no single correct resolution exists, a phenomenon rooted in \emph{value pluralism} \citep{Berlin1969FourEssays}. Human organizations have developed institutional safeguards over decades to navigate these tensions, such as frontline discretion, escalation procedures, audits, and precedent, as studied in street-level bureaucracy research \citep{Lipsky1980StreetLevel,nothdurfter2018meeting}.
LLM agents, however, may operate without comparable safeguards. When the governing policy is ambiguous, an agent may exploit discretionary space, overcommit to one interpretation, or produce a valid but unintended outcome while still appearing policy-compliant.
This is not hypothetical, as Claude Opus 4.5 recently resolved a $\tau^2$-bench \citep{barres2025tau} airline-booking task by exploiting a policy loophole, thus producing a technically valid but unintended resolution \citep{grace2026demystifying}.


    

These limitations expose a gap in current agent evaluation. Benchmarks such as $\tau$ and $\tau^2$ incorporate simplified domain policies and often rely on limited evaluation scopes, such as binary acceptability \citep{barres2025tau}. As a result, they provide limited insight into how agents reason under realistic policy ambiguity, where multiple interpretations may be defensible and outcomes must be evaluated along several dimensions.

To address this gap, we introduce \texttt{DRIP-R} (\textbf{D}ecision-making and \textbf{R}easoning \textbf{I}n ambiguous \textbf{P}olicy for \textbf{R}etail), a benchmark for evaluating LLM agents in dynamic conversational settings governed by realistic, ambiguous domain policies. Grounded in retail order returning scenarios, \texttt{DRIP-R} assesses not only whether agents reach acceptable outcomes, but also how they interpret policy ambiguity, justify decisions, and balance competing stakeholder interests.

As illustrated in Figure~\ref{fig:overall_benchmark}, we construct \texttt{DRIP-R} by first systematically identifying ambiguities in Amazon’s return policy. 
Based on these ambiguities, we design order-return tasks and present them to the agent in a conversational scenario. To evaluate agent performance, we develop an evaluation framework spanning policy adherence, behavioral alignment, conversational capabilities, and resolution quality. We evaluate \texttt{DRIP-R} across multiple LLMs and report a fine-grained analysis of agent behavior under ambiguity.
Our findings show that policy ambiguities systematically challenge LLM decision-making in agentic scenarios. Frontier models fundamentally disagree under ambiguity, and higher intra-model than inter-model agreement shows that each model interprets policy differently. Personas on both sides further shift outcomes, and customer personality predicts favorability, while agent disposition predicts denial rate. Extraverted and neurotic personas receive more favorable resolutions than agreeable ones.

Together, our work makes three contributions. First, we introduce \texttt{DRIP-R}, an open-source benchmark for evaluating LLM agents in realistic, policy-ambiguous retail return scenarios. Second, we develop a multidimensional evaluation framework and simulation infrastructure to assess agent behavior across policy adherence, behavioral alignment, conversational quality, and resolution accuracy. Third, we provide an empirical analysis across LLMs, ambiguity types, and agent personas, revealing fine-grained patterns in how agents reason, interact, and decide under such conditions.

\section{Preliminaries and Problem Formulation}
\label{sec:problem_formulation}

\subsection{Domain Policies and Ambiguities}
\label{subsec:domain_policies}

\citet{zwerdling2025towards} defines policies as documents containing domain-specific rules and constraints in natural language. Following \citet{crawford1995grammar}, we adopt a broader view of policies as \textit{rules, norms, and shared strategies} that structure situated action. This is particularly apt for retail, where eligibility depends on norms around customer treatment and discretionary judgment as much as on explicit rules. Resolution therefore requires reasoning over product condition, purchase timelines, customer circumstances, category-specific exceptions, and eligibility criteria. We ground \texttt{DRIP-R} in Amazon's publicly available return policy, which spans diverse product categories, conditions, and edge cases.

We define \textbf{policy ambiguity} as a property of a policy statement that permits more than one valid interpretation depending on context \citep{fowler2023strategies}. This definition aligns with broader accounts of ambiguity as the coexistence of multiple, potentially conflicting interpretations of situations or processes \citep{zahariadis2016delphic}. Under this definition, Amazon’s return policy offers a realistic basis for constructing tasks that test how agents interpret and act under ambiguous domain rules.





\subsection{Problem Setup}
\label{subsec:problem_setup}

In this section, we formalize the setting in which the LLM customer service agent must handle a user's return request in accordance with a domain policy that admits multiple defensible interpretations through a conversational setup. 

\paragraph{Domain policy.}

Formally, we define our benchmark policy as a sequence of natural-language statements that express these rules, norms, and shared strategies: $B_P = (p_1, p_2, \dots, p_N)$, where statement $p_i$ can admit a (possibly singleton) set of $k_i$ plausible interpretations $\mathcal{I}(p_i) = \{\iota_1, \dots, \iota_{k_i}\}$.

\paragraph{Tasks and outcomes.}
A task $\tau \in B_T$ is a user-centered return request with details on the purchased items and the purchase and delivery timeline. By design, each task $\tau$ depends on at least one ambiguous policy clause $p_i\in B_P^{\text{amb}}$, meaning that $p_i$ affects how $\tau$ should be resolved. $B_O$ is the set of all possible outcomes. In \texttt{DRIP-R} each outcome belongs to one of seven outcome categories: \textit{(1) Full Refund, (2) Partial Refund, (3) Refund with Gift Card, (4) Deny Refund, (5) Replacement, (6) Escalate to Human agent, (7) User Abort.}


\paragraph{Task ambiguity.}
Each of the tasks are paired with a finite set of \emph{plausible outcomes} $B_O(\tau) = \{o_1, \dots, o_{m_\tau}\}\subset B_O$, where each $o_j$ corresponds to a plausible interpretation $\iota_j \in \mathcal{I}(p_i)$ for some $p_i$ the task depends on.
A task is \emph{ambiguous} iff $|B_O(\tau)|\geq2$. Since all tasks in \texttt{DRIP-R} are ambiguous by construction, no $o_j$ is uniquely correct: $B_O(\tau)$ contains \emph{multiple} resolutions that are defensible under ambiguous policy statements.

\paragraph{Scenarios.}
A scenario is a pair $\sigma = (\tau, \pi)\in B_T\times\Pi$, where $\pi \in \Pi$ is a user persona drawn from a collection of personas (in our case, from PersonaHub dataset \cite{ge2024scaling}). The persona conditions the user simulator's behavior, communication style, and stated preferences. Tasks are constructed in a way such that for a scenario $(\tau, \pi)$, $B_O(\tau)$ does not depend on $\pi$.

\paragraph{Simulation environment.}
Given a benchmark scenario $\sigma$, a simulated user and customer service agent interact to resolve it. Our simulation framework follows the Dec-POMDP (Decentralized Partially Observable Markov Decision Process) formalism from $\tau^2$-bench. The benchmark environment for two-player simulations is defined as a tuple $\bigl(\mathcal{S},\,\{\mathcal{A}_i\},\,\{\mathcal{O}_i\},\,\mathcal{T},\,\mathcal{M},\,\mathcal{U}_\sigma,\,\mathcal{E}\bigr)$, with $i \in \{\text{agent},\,\text{user}\}$.
$\mathcal{S}$ denotes the state space, with dialogue histories and the underlying simulated database states; $\mathcal{A}_i$ denotes the action space of player $i$, with tool access restricted to the customer-service agent in our setting; and $\mathcal{O}_i$ denotes the observation space of player $i$, including incoming messages and tool responses. The transition function is defined as $\mathcal{T}: \mathcal{S} \times \mathcal{A} \rightarrow \mathcal{S} \times \mathcal{O}$. The message space $\mathcal{M}$ contains all natural-language messages, and $\mathcal{U}_{\sigma}$ denotes the instruction space for scenario $\sigma$, with both general domain policies and task-specific constraints. Finally, $\mathcal{E}$ is the multidimensional evaluation function defined below, replacing the binary reward function $\mathcal{R}$ used in $\tau^{2}$-bench.

\paragraph{Conversations.}
The outcome of the simulation of a benchmark scenario $\sigma$ is a conversation $\mathcal{C}=(T_1, T_2, \dots, T_n)$, where each turn $T_i=(t_1, t_2, \dots, t_m)$, with $t_i\in\mathcal{A}_{agent}\cup\mathcal{A}_{user}$, is a possibly empty sequence of tool calls followed by a single message $t_m$ from the user or agent to their counterpart. In \texttt{DRIP-R} only the agent is able to make tool calls. The conversation ends either because $T_n$ contains an end-of-conversation message from the agent or user, or because $n$ reaches the maximal turn limit $t_{lim}=20$.

\paragraph{Resolution function.}

The \emph{resolution function} $\rho: \mathcal{C}\rightarrow (B_O \cup \{\bot\})\times \mathcal{R}$ maps a completed conversation to a resolution $\rho(\mathcal{C})=(o, r)$, where $o$ is the outcome the agent committed to, or $\bot$ if the conversation ended due to user abort or reaching the turn limit, and $r\in\mathcal{R}$ is the reason produced by the agent model at the conclusion of the conversation (if $o=\bot$, then $r=\bot$). In \texttt{DRIP-R}, $\rho$ is the customer service agent model: it generates a final outcome statement and reason based on the conversation when it resolves a return request.



\paragraph{Evaluation.}

We replace the binary reward function in $\tau^2$-bench with a multidimensional evaluation function $\mathcal{E}(\tau, \mathcal{C}, \rho(\mathcal{C}))=e\in\mathbb{R}^d$ over $d$ axes (policy adherence, dialogue quality, behavioral alignment, interest alignment, task-resolution alignment, and their sub-dimensions) that evaluates a conversation in the context of the task and the resolution reached by the agent. Therefore, $\mathcal{E}$ does not yield a binary success signal, but a ground for comparing agents across multiple evaluation dimensions, as defined in Section \ref{subsec:eval_pipeline}.



\section{Benchmark Construction}
\label{sec:benchmark_construction}


\subsection{Dataset Construction}
\label{subsec:dataset_construction}


The dataset construction process begins with identifying and extracting notable ambiguities from the policy document. Using \citet{massey2014identifying} and \citet{gervasi2019ambiguity} as the reference taxonomy for ambiguity identification, we prompt an LLM (GPT-4.1) to identify ambiguities in the policy. The identified ambiguities are then filtered and verified by two human experts (IAA of 84.62\%), who validate each ambiguity identified by the LLM against the policy. An example of the ambiguities identified, along with relevant policy clauses, is shown in Table \ref{tab:example_policy_ambiguities} of Appendix \ref{app:ambiguity_types}, which also provides definitions for each ambiguity type.   

\begin{figure*}[!htbp]
    \centering
    \includegraphics[width=1.0\textwidth]{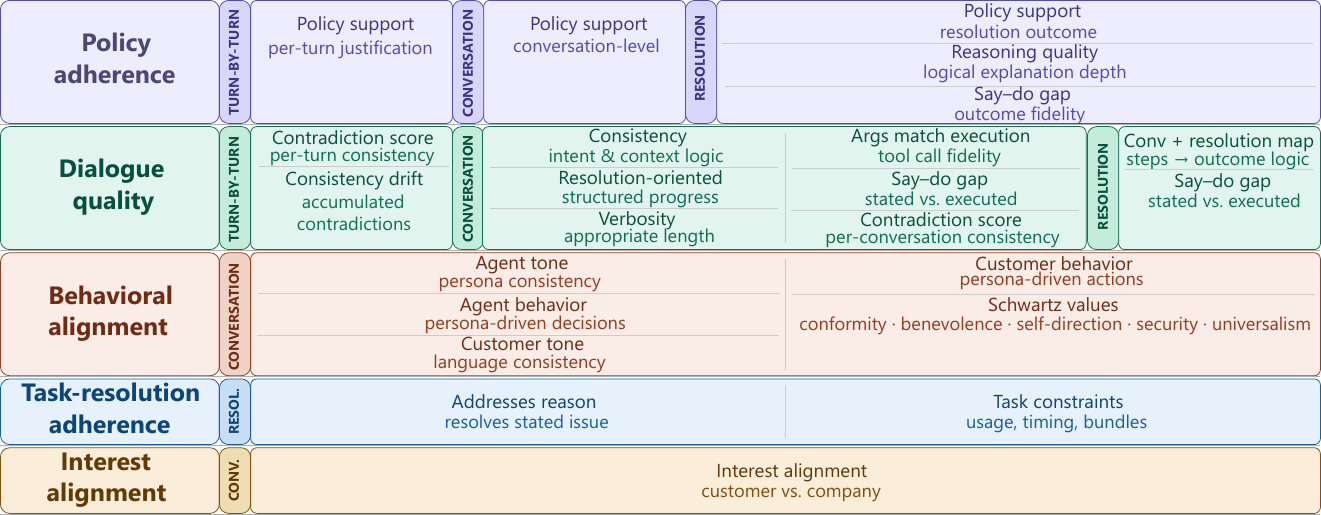}
    \caption{
    Overview of the evaluation dimensions. The evaluation framework includes five dimensions, each with multiple sub-metrics evaluated at the turn/conversation/resolution level.}
    
    
    \label{fig:evaluation_framework}
\end{figure*}

The primary task we designed for this dataset is Order Return, where the customer wants to return an order placed on an E-commerce site and negotiate its resolution with a service agent. For product details, we use an open-source Amazon product description dataset obtained from HuggingFace\footnote{\href{https://huggingface.co/datasets/philschmid/amazon-product-descriptions-vlm}{Amazon product description dataset at HuggingFace}}. 

To create each task, we randomly sample two or three items from the Amazon product description dataset (ensuring no combination is reused across tasks) and pass them, along with the extracted ambiguities, to GPT-4o-mini. The model is prompted to generate a realistic Amazon return scenario that depends on at least one specific ambiguity whose resolution changes the outcome, that elicits multi-turn customer-agent dialogue, and specifies the order timeline, item-level conditions, delivery status, and customer circumstances. 

Using these criteria, we collect an initial pool of 500 tasks, which are subsequently stratified, validated, and sampled down (through stratified sampling) to 40 high-quality tasks, as described in Appendix \ref{app:task_diversity}. The tasks cover all the ambiguity types in our taxonomy and produce resolutions across all seven outcome categories (per-outcome task statistics are provided in Appendix \ref{app:dataset_stats}). We use stratified sampling for the following reasons: 1) to maximize diversity along four axes, 2) to cover diverse situational setups, 3) to ensure policy-coverage breadth, 4) to maintain product-domain variety and condition-ambiguity depth. This sampling yields nine diversity-by-ambiguity-type strata, with a complexity-diversity separation of $\rho$ = 0.743. No product combination is reused across tasks. Each of the final tasks was manually verified based on whether it could realistically occur 
in a real-world scenario. Additionally, for 10 tasks, two human annotators annotated which outcomes are plausible for each task, as described in Appendix \ref{app:task_outcome_annotation}. We found substantial disagreement between human annotators, further confirming the ambiguity of our tasks. Each task is created with multiple ambiguity types, with a mean of 2.95 ambiguities per task. The most frequent pairing of ambiguity type is Vagueness with Coordination Ambiguity.

For customer persona creation, we augment (using GPT-4o-mini) elite personas from Persona-Hub\footnote{\href{https://huggingface.co/datasets/proj-persona/PersonaHub}{Persona-Hub dataset at HuggingFace}} with personality (Big Five \citep{big5personality}), communication style (rapport/report-oriented, direct/indirect, linear/non-linear), language traits (Joos's \textit{Five Clocks} \citep{zgusta1966martin}), and demographic attributes (age, purchase history, income range), which collectively condition the customer simulator's behavior. From an initial pool of 30, 10 personas are retained through manual selection for realism and coverage. We pair each persona with each task, generating $10\times 40 = 400$ scenarios (statistics in Appendix \ref{app:dataset_stats}).





\subsection{Evaluation Dimensions}
\label{eval_framework}

Our evaluation takes as input the task, the simulated conversation, and the agent's resolution. We evaluate three levels (\emph{turn}, \emph{conversation}, \emph{resolution}) along five dimensions with multiple subdimensions (Figure \ref{fig:evaluation_framework}; full details in Appendix \ref{app:eval_dimensions}). Dimensions are designed to be largely independent, so an agent trace can score well on one and poorly on another.
Below we describe the five top-level dimensions, with detail on sub-dimensions used in our analysis.
    \paragraph{Policy adherence} defines the extent to which the agent's messages at each turn and across the conversation, as well as the final resolution, are grounded in the provided policy text. It assesses justifiability of the agent's policy usage, not correctness. This matters under ambiguity, where multiple defensible resolutions exist.
    

    \paragraph{Dialogue quality} captures the agent's conversational competence, independent of policy or persona. The sub-metrics assess whether the conversation is coherent, focused, and faithful to its own commitments.
    

    \paragraph{Behavioral alignment} measures the extent to which the agent and simulated user act consistently with their assigned personas. We use six values from Schwartz value theory \citep{schwartz2012overview}: \emph{Conformity}, \emph{Benevolence}, \emph{Self-Direction}, \emph{Security}, \emph{Universalism}, and \emph{Power}. These ratings capture value priorities revealed by the agent's conduct across the conversation.

    


    \paragraph{Task-resolution adherence} measures whether the resolution addresses the user's specific request, independent of policy. This dimension distinguishes resolutions that are nominally permitted by policy from those that actually fit the task at hand.
    


    \paragraph{Interest alignment} captures how well the conversation and resolution align with customer and company goals. Company interest alignment is 
    a policy-compliance score (1–5) for the company's benefit. Customer goal alignment is measured in two steps: customer goals extracted from task description and each goal is individually assessed for overall satisfaction ratio. 

\begin{figure}[t]
    \centering

    \begin{subfigure}[t]{\columnwidth}
        \centering
        \includegraphics[width=\linewidth]{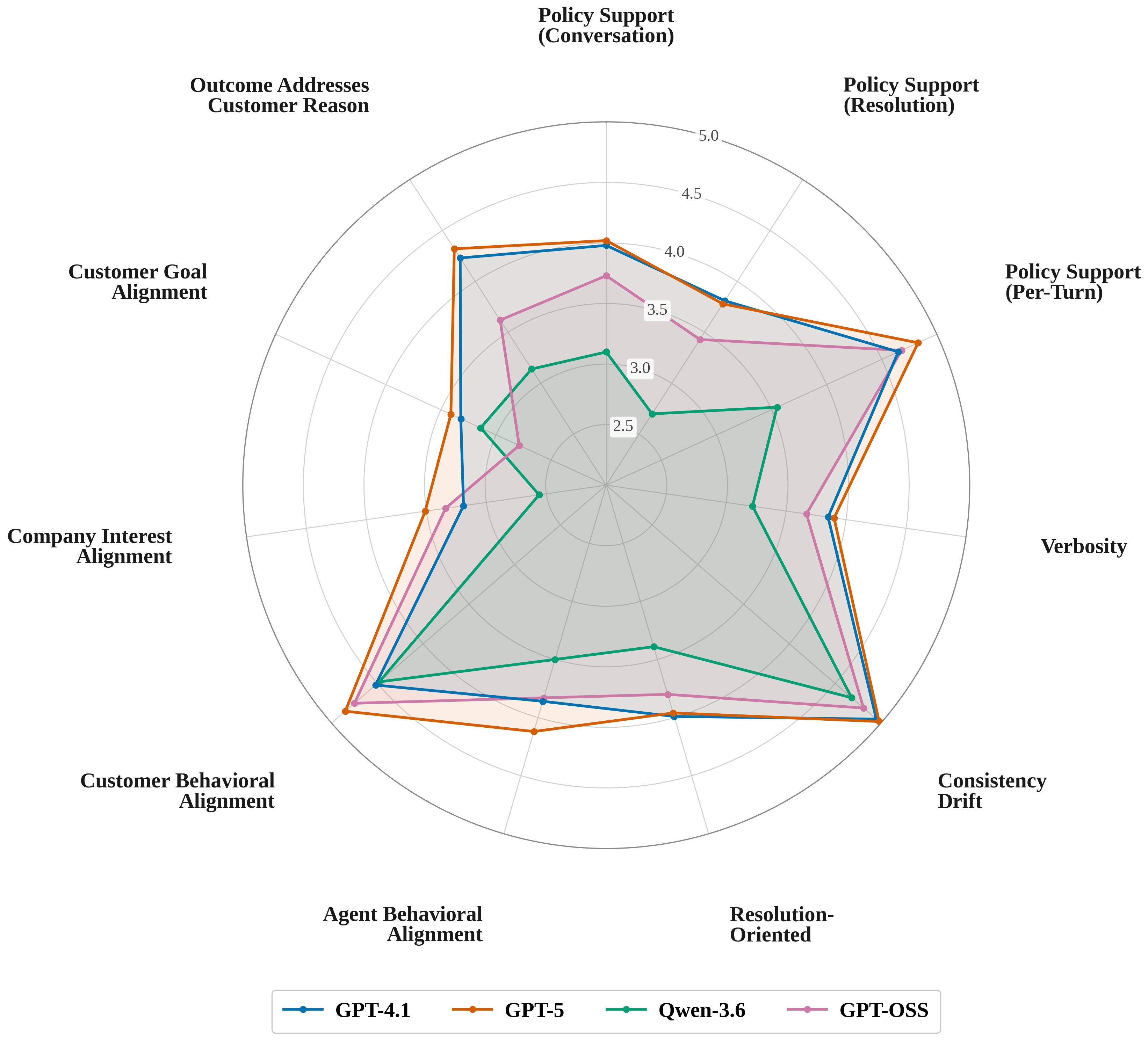}
        \caption{
        Overall model performance comparison across the evaluation metrics.
        }
        \label{fig:all_performance_spider}
    \end{subfigure}
    \hfill
    \begin{subfigure}[t]{\columnwidth} 
        \centering
        \includegraphics[width=\linewidth]{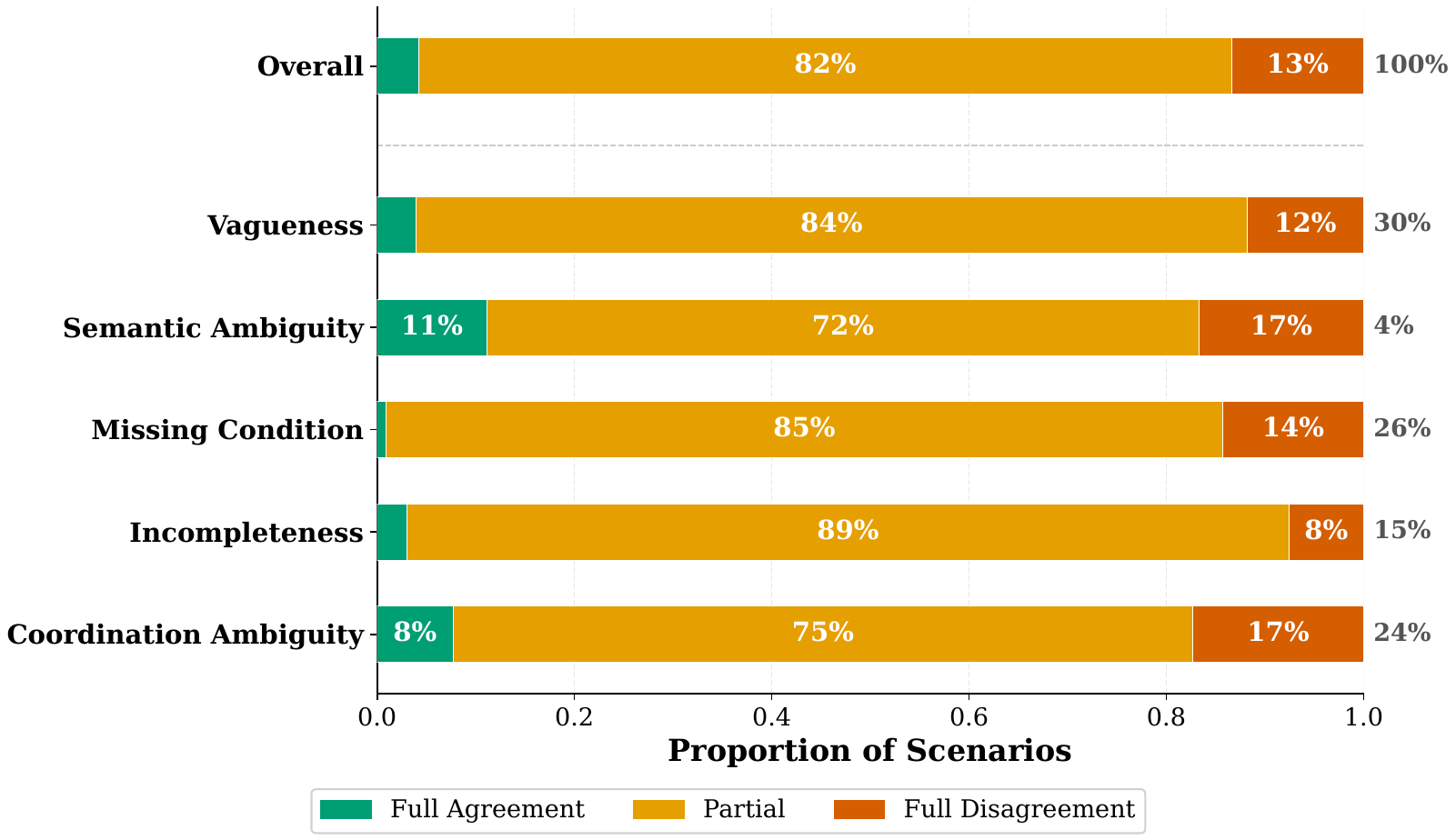} 
        \caption{
        Cross-model resolution agreement by ambiguity type across four models. 
        }
        \label{fig:cross_model_res_agreement}
    \end{subfigure}

    \caption{
    Results of overall model performance and cross-model resolution agreement.
    }
    \label{fig:model_performance_and_agreement}
\end{figure}

\subsection{Evaluation Pipeline}
\label{subsec:eval_pipeline}


We implement the evaluation framework using the LLM-as-a-Judge approach \citep{gu2024survey}. We use parallel judges, each focusing on a particular part of the evaluation (Policy Adherence + Task-Resolution Adherence, Dialogue Quality, Behavioral Alignment, and Interest Alignment), plus an orchestrator that validates each judge's output format and aggregates the results into a single result file. This reduces the load on each judge's context window and allows us to calibrate the LLM judges independently as described below. All judges use \texttt{gpt-5.4-2026-03-05} with temperature 0 to ensure deterministic outputs. Two independent-lab judges reproduce model rankings (which we discuss in detail in Appendix \ref{app:cross-judge-agreement}).

\paragraph{LLM-as-Judge calibration.}
Calibration of judges for each dimension per human preference is essential to maintain judge quality \citep{tan2024judgebench}. We design a multi-stage calibration process grounded in human annotations collected per evaluation metric. The process starts by collecting human preferences through annotation. Human annotations are collected for a representative sample of 30 instances (each being a task, conversation, and resolution) and include metrics from our Judge pipeline (as described in Appendix~\ref{app:human_annotation}). Each of the three expert annotators labels 15 conversations, yielding a pair of annotators assigned disjoint sets of 5 conversations each. The pairwise annotation results are shown in Table~\ref{tab:annotator_agreement} in the appendix, with a mean exact agreement of 0.23. However, most weighted Cohen's kappa ($\kappa_w$, quadratic weights) and Krippendorff's alpha ($\alpha_K$, ordinal distance) values are at or below chance, consistent with the task being genuinely ambiguous for human annotators at the 5-point granularity. Additionally, the higher adjacent (within $\pm 1$) agreement of 0.72 reinforces this reading, indicating that annotators converge on the direction of a judgment but differ on the specific integer point on the scale. Finally, the ground truth for calibration was established through adjudicated consensus, with an explicit consensus label recorded based on the discussion between two annotators and an adjudicator.



For calibration, we use two approaches: Distribution Preserving Few-Shot sampling (DPFS) \citep{jang2025instajudge} and G-Eval \citep{liu2023g}. 
We take 5 DPFS-selected samples (out of 30) for calibration and few-shot prompting. The remaining 25 form a held-out test set for LLM-as-judge evaluation. Since our metrics do not rely on value correctness, we focus on minimizing bias (difference between average human and LLM scores) and maximizing adjacent agreement (within-one-point match between human and LLM) \citep{selvakumar2025multivox}. The calibration results along with the respective metric definitions are reported in Table~\ref{tab:agreement_metrics_by_dimension} (Appendix~\ref{app:judge_calibration}). The mean bias of the combined approach is $+0.33$, with adjacent agreement of 0.84, suggesting that the judge agrees with human annotators within one point on 84\% of 1--5 ordinal scores and preserves their ordering. We report the signed per-metric bias in Table ~\ref{tab:annotator_agreement}, which ranges from $-0.29$ (Reasoning Quality) to $+0.88$ (Resolution Description Fidelity). The table reports comparative quantities (model vs. model, control vs. no-control, persona vs. persona, type vs. type). The Policy Support metrics for policy adherence show near-zero bias (+0.04 for conversation and +0.08 for resolution), and resolution-type classification reaches 56\% accuracy on the 7-class problem. 

\section{Experiment}
\label{sec:expt}




\begin{figure*}[!h]
    \centering
    \includegraphics[width=\textwidth,
        height=0.4\textheight,
        keepaspectratio]{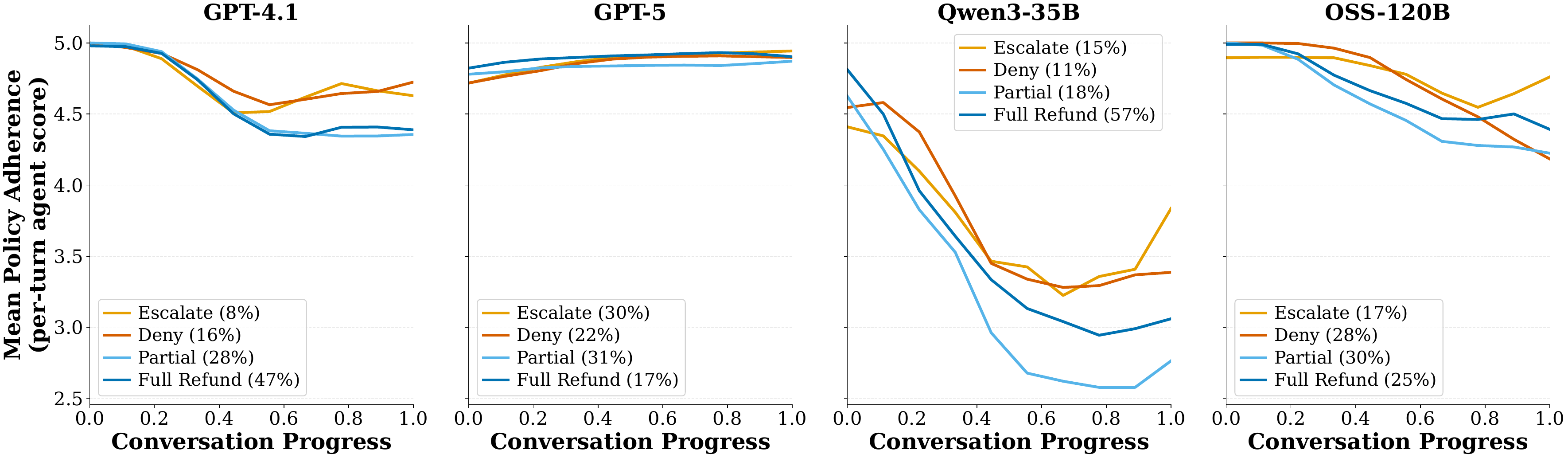}
    \caption{
    Model-wise policy adherence trajectory by resolution. 
    X-axis shows normalized conversation progress; Y-axis shows mean policy-adherence scores per turn.
    }
    \label{fig:trajectory_analysis}
\end{figure*}


\subsection{Experiment Setup}
\label{subsec:expt_setup}

We evaluate four models selected to reflect recent LLM developments and their suitability for customer-facing agentic tasks: two open-source (\texttt{gpt-oss-120b}, \texttt{Qwen3.6-35B-A3B}) and two API-only (\texttt{gpt-5-2025-08-07}, \texttt{gpt-4.1-2025-04-14}). 
We use \texttt{gpt-4o-mini} as the user simulator to provide consistent, cost-efficient customer turns. When evaluating the customer service agent in the benchmark, we hold the user simulator constant across all agent models and do not evaluate it.
Outputs of all scenarios are evaluated via the LLM-as-a-Judge pipeline (Section~\ref{subsec:eval_pipeline}) using \texttt{gpt-5.4-2026-03-05} judge. Agent and customer models use temperature 0.7 to ensure response diversity.

 

Each model receives a scenario-specific description containing only the information relevant to its role. Full task details, including policy ambiguities and return specifics, are withheld to reflect realistic deployment conditions. Every turn must contain a message, preceded by a sequence of tool calls from the customer service agent, whose prompt instructs it to collect all relevant facts before making a decision. We describe the full simulation process with an illustrative example in Appendix~\ref{app:conv_simulation}. 

\subsection{Results and Analysis}
\label{sec:results}

We present results for a selected set of evaluation dimensions across all models in  Figure \ref{fig:all_performance_spider}, with additional details provided in  Table \ref{tab:main_results} in the appendix. 
The scores are the average Likert-scale rating for each metric per model. 
From Figure \ref{fig:all_performance_spider}, it is clearly observed that GPT-5 shows better performance on in-conversation metrics. The two exceptions are at the resolution level, where GPT-4.1 narrowly edges GPT-5 on Policy Support (Resolution) and on Resolution-Oriented Responses. Aggregate metric means are stable across independent runs and scenario resamples (Appendix \ref{app:subsec:run_to_run}); the largest standard deviation on any major evaluation dimension is $\pm0.11$, with model orderings preserved.





\begin{figure*}[t]
    \centering
    \includegraphics[width=\textwidth,
        height=\textheight,
        keepaspectratio]{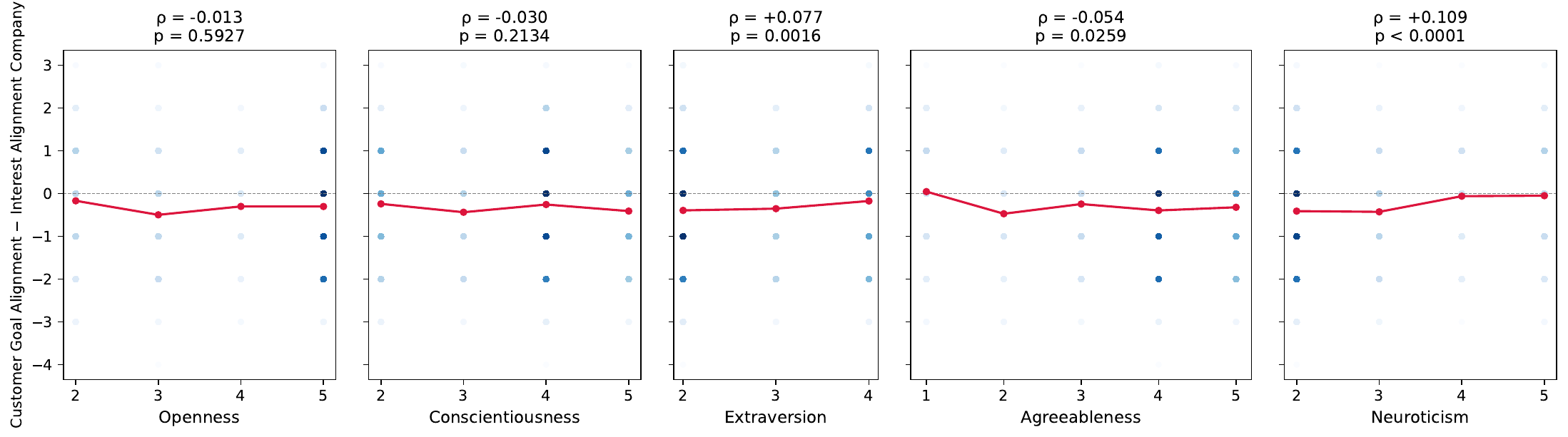}
    \caption{Alignment balance (Customer Goal $-$ Company Interest Alignment) across Big Five trait levels. Dot color: point density (darker = denser); red lines: group means per trait score. Spearman $\rho$ and $p$-values (top) test monotonic association.
    }
    \label{fig:user_persona_alignment}
\end{figure*}

\begin{figure*}[!h]
    \centering
    \includegraphics[
        width=\textwidth,
        height=0.7\textheight,
        keepaspectratio
    ]{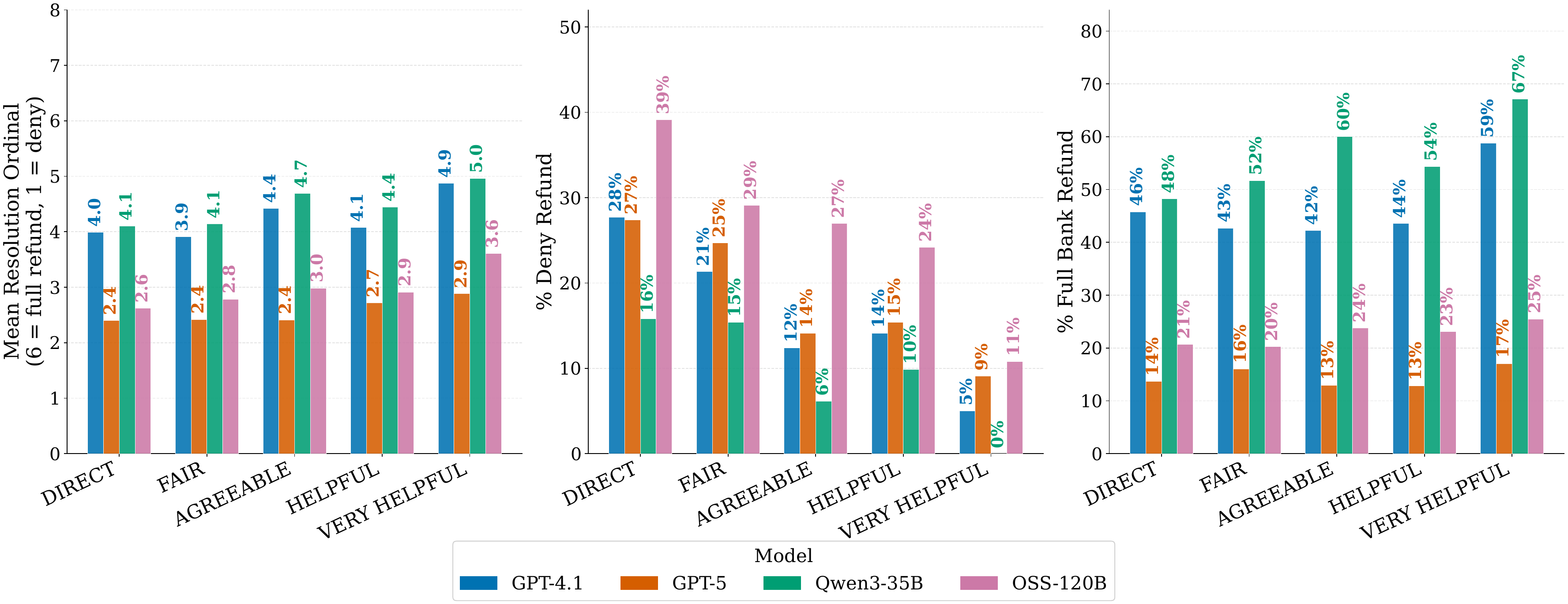}
    \vspace{-0.5em}
    \caption{
    Cross-model resolution signals by agent persona. Mean Resolution Ordinal ranks customer-favourability:
    Abort=0, Deny=1, Escalate=2, Gift Card=3, Partial=4, Replacement=5, Full=6.
    }
    \label{fig:cross_model_res_persona}
    \vspace{-0.8em}
\end{figure*}


\paragraph{Impact of policy ambiguity on model resolution.}
We inspect resolution instability across both models and repeated runs to demonstrate the impact of policy ambiguity. As a cross-model measure, we classify each scenario's outcomes as full agreement (one outcome across all four models), partial agreement (2–3 distinct outcomes), or full disagreement (four distinct outcomes), and report the breakdown by ambiguity type in Figure \ref{fig:cross_model_res_agreement}.
Overall, only 4.2\% of scenarios yield full agreement (all four models converge on the same resolution), and 13.3\% yield full disagreement (every model produces a distinct outcome). The remaining 82.4\% show partial agreement (some but not all models converge). 

To contextualize these results, we ran each model three times on a stratified subsample (appendix~\ref{app:stability_analysis}). Mean pairwise intra-model agreement ranged from 28.6\% for GPT-OSS-120B to 69.5\% for GPT-4.1 (Table~\ref{tab:intra_model_agreement} in the appendix). 
While these intra-model agreement rates are substantially higher than the 4.2\% full cross-model agreement rate, they also show that repeated runs of the same model can still produce different resolutions. Notably, although three of the four models originate from OpenAI, we verify that intra-OpenAI agreement is not systematically higher than cross-lab agreement (Appendix \ref{app:stability_analysis}), indicating that the models from the same source do not confound our cross-model findings. Thus, the low cross-model full-agreement rate reflects both systematic differences between models and broader resolution instability under ambiguous policy language. In addition, we provide a policy-controlled ablation (Appendix \ref{app:ctrl_baseline}) where the agent receives a deterministic rule for each ambiguity, raising Policy Adherence by +0.26 without affecting Behavioral Alignment, showing that resolving ambiguity directly improves outcome quality. 

Examining the behavior-decision gap (conversation-level minus resolution-level Policy Adherence) across ambiguity types reveals that Coordination Ambiguity is the hardest for agents to resolve consistently with their stated reasoning (mean gap 0.29, significantly higher than three of four other types of ambiguities). We  discuss this finding further in Appendix \ref{app:subsec:policy_shaping}.

\paragraph{Impact of ambiguities on conversations.} 
To identify the impact of policy ambiguities on conversations, we analyse conversation trajectories to determine how ambiguity manifests at the conversational level and whether the resulting decisions are 
internally consistent with the agent's per-turn reasoning. We trace the mean per-turn Policy Support score, grouped by resolution (\textsc{ABORT} excluded). Conversations of varying length are normalised to a relative position axis via interpolation and then 
averaged within groups (Figure~\ref{fig:trajectory_analysis}). Across all four models, escalation maintains end-of-conversation adherence at or near the top of the bucket ordering, while refund-granting outcomes drop further than escalation from their early-conversation levels. The drift is most striking in Qwen3-35B and OSS-120B, where buckets diverge around the conversation midpoint. In contrast, GPT-4.1 shows a smaller, earlier dip that flattens, and GPT-5 stays uniformly high ($\geq 4.83$) across outcomes.

\paragraph{Impact of user personas on alignment.}
We tested, whether different types of user personas systematically produce different kinds of outcomes in return scenarios. Specifically, we looked at the relationship between the personality of the user persona and the balanced alignment score (customer goal alignment - interest alignment company). Figure \ref{fig:user_persona_alignment} shows the group means for each personality trait level. We report Spearman correlations to account for ordinal personality trait data. For Openness and Conscientiousness we found no significant correlation. Though effect sizes were small overall, for Extraversion and Neuroticism we found that highly extraverted ($\rho=0.077, p=0.0016$) and highly neurotic ($\rho=0.109, p<0.0001$) user personas were associated with results that favor customer goal alignment over company interest alignment. Whereas user personas that were more agreeable were associated with lower customer goal alignment scores and higher company interest alignment scores ($\rho=-0.054, p=0.0259$). This raises questions about the fairness of customer service agents, which currently appear to reward extraverted, neurotic and less agreeable customers.


\paragraph{Impact of agent personas on outcomes.}
To examine how agent personas shape outcomes, we map outcomes to an ordinal scale (0-6, with higher values indicating more customer-favorable outcomes) and compute, for each agent, the mean outcome score across models (Figure~\ref{fig:cross_model_res_persona}). In the Figure, the ordinal is a single-stakeholder measurement axis capturing how favorable an outcome is to the customer. Ordering outcomes by customer benefit asserts nothing about which is more correct (or acceptable), as the correct outcome may not always be customer beneficial. For example, a Full Refund is simply more customer-favorable than a denial from the customer's standpoint.
Personas reliably anchor the final outcomes: \textsc{Direct} agents consistently yield the highest denial rates, whereas \textsc{Very\_Helpful} agents yield the lowest across all models. However, what replaces denials (full/partial refund, or escalation) depends on the model. The association between the agent persona and the denial rate is statistically significant in the pooled analysis ($\chi^2(4) =59.03$, $p < 0.001$, Cramér`s $V = 0.19$), and is significant independently in each of the four models (per-model $p = .001 - .008$). This confirms that \textsc{Very\_Helpful} agents deny requests significantly less often than \textsc{Direct} agents. 




\section{Conclusion}
\label{sec:conclusion}
We introduce \texttt{DRIP-R}, a benchmark for studying LLM decision-making under domain policy ambiguity in retail. We construct a dataset of real-world tasks and customer personas to evaluate LLM agents in a dual-control conversational setup, paired with a multi-judge framework spanning policy adherence, dialogue quality, and resolution alignment. Experiments across models reveal how policy ambiguities affect LLM behavior, conversational trajectories, and resolutions. \texttt{DRIP-R} establishes a foundation for accountable AI systems that operate under the inherent ambiguity of real-world policy and offers insights into the underexplored dimension of agentic behavior under such conditions. As a first step toward exploring diversity in domains, we adapt DRIP-R to the airline domain (Appendix \ref{app:cross_domain_airline}), where the construction and simulation pipeline transfers unchanged and reproduces the core disagreement effect. Future work will extend \texttt{DRIP-R} to other high-stakes domains with ambiguous policies (e.g.,
social welfare, healthcare, insurance, airline customer service) and explore governance mechanisms that make agent discretion transparent, verifiable, and auditable.

\section*{Limitations}
\label{sec:limitations}
We acknowledge the limitations of this work. First, the benchmark currently relies on a single primary policy document and one related policy document; incorporating a broader set of real-world policies would improve the findings' generalizability. The benchmark contains 40 tasks and 10 personas, which is small for the number of tasks. Pairing them generates 400 total scenarios, and resolving each with four agent models produces 1,600 agent-customer conversations, supplemented by 1,176 intra-model stability rollouts (Appendix C.3) and a policy-controlled ablation (Appendix C.1). Table \ref{tab:benchmark_scenario_counts} places this scenario count alongside comparable agent benchmarks. Extending the benchmark is part of our future work, since our proposed construction framework can easily generalize to additional domains, policies, tasks, and corresponding scenarios. Second, testing the benchmark on a larger set of models will draw a better picture of how LLM agents operate under policy ambiguity and will strengthen our findings. 

Another limitation is the use of LLMs at several stages of the benchmark construction. Future versions can reduce LLM dependency at each stage by using formal underspecification detection and expert-oriented decisions to extract ambiguity. For scenario construction, real-world return cases and templates can be used; real behavioral data and human role-play for personas; and expanded human annotation for evaluation. However, this future work aims not to remove LLMs but to shift from generate-then-verify toward human and real-data authoring with LLM assistance. 
Finally, comparing LLM resolutions against decisions made by human customer-service representatives on identical scenarios would clarify whether the observed disagreement patterns are LLM-specific or reflect inherent task ambiguity. Appendix I.1 provides model-independent evidence on this point. In the appendix, we show that the annotators saw only the task and the policies, never a model output, and diverged as strongly as the models did. Their disagreement therefore indicates that the ambiguity is intrinsic to the tasks, and the annotations approximate how human representatives might resolve the same scenarios. We make no claim about how humans and LLMs differ, and frame all findings as insights into how LLMs handle policy ambiguity in customer return tasks. Related benchmarks such as $\tau$-bench and $\tau^2$-bench likewise omit human baselines. A full resolution-level comparison remains valuable, and we take it as our next step.







\section*{Acknowledgment}
This work was funded by the RobustLLMAgents project in partnership with Amazon Science, Berlin. We acknowledge our collaborators at Amazon for their invaluable support. We also acknowledge Minh Vu Pham and Tewodros Bizuneh for annotations. Finally, we thank the members of the IT:U NLP group for helpful suggestions that improved the project at various stages.

\bibliography{custom}

\appendix
\section{Related Work}
\label{sec:related_work}
\paragraph{Policy ambiguity in real-world domains.}
Policy ambiguity refers to the existence of multiple valid interpretations of clauses or phrases within a policy, arising from vagueness or under-specification in the policy language \citep{zahariadis2023ambiguity}. Such ambiguities manifest in case-specific scenarios, where differing interpretations of the same policy content lead to divergent outcomes, causing goal displacement in regulatory practice \citep{huizinga2021exploring}.  These interpretive differences are task-dependent, as the impact of a given ambiguity varies with the context in which the policy is applied \citep{matland1995synthesizing, bhaskar2025exploring}. Their impact is double-edged: ambiguities can facilitate task resolution by allowing flexibility \citep{ang2024ambiguity,qi2025toward} while simultaneously complicating it through misalignment of interpretations \citep{zahariadis2016delphic}. Furthermore, ambiguities are actor-specific, as vagueness within a policy allows individual actors to impose their own interpretation \citep{fowler2023strategies}. Taken together, policy ambiguities are an inherent feature of real-world domain policies and, irrespective of intent, they shape task-oriented decision-making in practice \citep{taylor2021ambiguity, leong2022policy}.

\begin{table*}[h]
\centering
\small
\begin{tabular}{lccr}
\toprule
\textbf{Benchmark} & \textbf{Real-World Policy} & \textbf{Document Structure} & \textbf{FK-Grade} \\
\midrule
$\tau$-bench \citep{barres2024tau}         & No & Single document & 9.55 \\
$\tau^2$-bench \citep{barres2025tau}       & No & Single document & 9.50 \\
ST-WebAgentBench \citep{levy2024st}        & Partial & Per-task rule instances & 7.94 \\
\bottomrule
\texttt{DRIP-R} (ours)                   & \textbf{Yes} & Multi-document & \textbf{11.89}\\
\bottomrule
\end{tabular}
\vspace{0.3em}
\caption{Comparison of domain policy usage across agentic benchmarks. FK-Grade reports the Flesch–Kincaid grade level of the policy document. Scores of 6–8 correspond to easy, middle-school-level text; 9–10 to fairly easy, high-school-level text; and 11–12 to moderate, college-level text.
}
\label{tab:policy_comparison}
\end{table*}

\begin{table}[t]
\centering

\setlength{\tabcolsep}{4pt}
\small
\begin{tabular}{@{}llr@{}}
\toprule
\textbf{Benchmark}
& \textbf{Domain}
& \textbf{\# Scenarios} \\
\midrule
$\tau$-bench / $\tau^2$-bench & Retail     & 115 \\
$\tau$-bench / $\tau^2$-bench & Airline    & 50  \\
$\tau^2$-bench                & Telecom    & 114 \\
ST-WebAgentBench              & Web agents & 375 \\
DRIP-R (ours)                 & Retail     & 400 \\
\bottomrule
\end{tabular}

\caption{Number of scenarios on which experimental results are reported.}
\label{tab:benchmark_scenario_counts}
\end{table}


\paragraph{LLM agents under policy ambiguities.}

In the literature, domain policies have seen growing use as a governance mechanism for agentic task execution. The $\tau$ series \citep{barres2024tau, barres2025tau, ray2026tau} benchmark conversational and voice agents with tool-calling capabilities against domain policies. Policy constraints have similarly been used as countermeasures in safety-oriented benchmarks such as Agent-SafetyBench \citep{zhang2024agent}, DoomArena \citep{boisvert2025doomarena}, and ST-WebAgentBench \citep{levy2024st}. Closer to our focus, \citet{richardson2025should} proposes an approach combining deliberation over shared meanings and authoritative prescriptions with practical policy use, while \citet{zieli2026operationalizing} decomposes policy documents into atomic obligations and flags under-specified or unverifiable requirements to guide policy refinement. However, the policies used across these benchmarks are synthetically constructed, simplified proxies rather than real-world policy documents, i.e., typically short, single-document specifications written in clean, unambiguous language to define a fixed set of rules for the agent to follow. As a result, they do not reflect the linguistic complexity, structural depth, or interpretive ambiguity that characterize policies deployed in practice. Table~\ref{tab:policy_comparison} quantifies this gap across three dimensions: (1) whether the benchmark uses a real-world policy, (2) the document structure, and (3) the Flesch--Kincaid (FK) readability score \citep{kincaid1975derivation}, where higher scores indicate higher reading grade level (i.e., longer sentences and more polysyllabic vocabulary). \texttt{DRIP-R} is the only benchmark grounded in a real-world, multi-document policy, with an FK score (11.89) substantially higher than prior benchmarks (7.94--9.55), reflecting the policy complexity that agents must navigate. Consequently, how agents behave when presented with tasks involving multiple policy ambiguities remains largely unexplored.

\section{Dataset Statistics}
\label{app:dataset_stats}
 Table \ref{tab:scenario_summary} provides a high-level overview and key statistics for the dataset. We also report the benchmark's coverage along the following three axes to characterize its representativeness within the grounding policy: the ambiguity types exercised (Table \ref{tab:ambiguity_type_distribution}), the resolution outcomes reached, and the task-complexity distribution (Very high: 57.5\%, High: 40\%, and Medium: 2.5\%). Another statistic from Table \ref{tab:scenario_summary} is that the persona pool is imbalanced by gender, with 280 female and 120 male scenarios. Gender is a demographic attribute of the personas rather than an analyzed fairness axis, since the fairness results in Figures 5 and 6 turn on Big Five traits and agent disposition. We find no gender disparity in outcomes. Mean customer favorability is nearly equal at 3.51 for female and 3.64 for male personas (Mann-Whitney p = 0.19), and denial rates are essentially identical at 17.6\% and 17.4\% ($\chi^2$ p = 0.97). While the imbalance does not affect the findings, a more gender-balanced persona set is planned for a future version of the benchmark.

    \begin{table}[t]
    \centering
    \small
    \setlength{\tabcolsep}{4pt}
    \renewcommand{\arraystretch}{1.08}

    \begin{tabularx}{\columnwidth}{@{}lXr@{}}
        \toprule
        \textbf{Category} & \textbf{Value} & \textbf{N} \\
        \midrule
        \multirow{3}{*}{Overview}
            & Total scenarios & 400 \\
            & Unique tasks & 40 \\
            & Unique personas & 10 \\
        \midrule
        \multirow{3}{*}{Task Complexity}
            & Very High & 230 \\
            & High & 160 \\
            & Medium & 10 \\
        \midrule
        \multirow{2}{*}{Persona Gender}
            & Female & 280 \\
            & Male & 120 \\
        \midrule
        \multirow{2}{*}{Text Length (mean char)}
            & Task description & 1,039 \\
            & First customer message & 314 \\
        \bottomrule
    \end{tabularx}

    \vspace{0.4em}
    \caption{Summary statistics of the scenario dataset.}
    \label{tab:scenario_summary}
\end{table}

\begin{table}[ht]
\centering

\setlength{\tabcolsep}{4pt}
\small
\begin{tabular}{lcc}
\toprule
\textbf{Ambiguity Type}
& \textbf{\% of Tasks}
& \textbf{\% Primary} \\
\midrule
Vagueness                 & 85.0\% & 30.0\% \\
Coordination Ambiguity    & 85.0\% & 25.0\% \\
Incompleteness            & 55.0\% & 15.0\% \\
Missing Condition         & 52.5\% & 25.0\% \\
Semantic Ambiguity        & 17.5\% & 5.0\%  \\
\bottomrule
\end{tabular}
\caption{Distribution of ambiguity types across tasks.}
\label{tab:ambiguity_type_distribution}
\end{table}

\begin{table}[t]
\centering

\setlength{\tabcolsep}{6pt}
\small
\begin{tabular}{lc}
\toprule
\textbf{Outcome} & \textbf{Share} \\
\midrule
Full Refund             & 34\% \\
Deny Refund             & 18\% \\
Escalate to Human       & 16\% \\
Partial Refund          & 13\% \\
User Abort              & 8\%  \\
Replacement / Exchange  & 6\%  \\
\bottomrule
\end{tabular}
\caption{Outcome-space coverage. Resolution categories reached when outcomes are pooled across the four evaluated models.}
\label{tab:outcome_space_coverage}
\end{table}

\section{Additional Experiments}
\label{app:additional_expts}

\subsection{Policy-controlled Setup}
\label{app:ctrl_baseline}

We design a policy-level control baseline that isolates the effect of hard-binding resolution behavior from that of free-agent judgment. The same LLM agent runs on the same 80 scenarios (20\% of the scenarios, selected through ambiguity stratification) thrice: (1) A disambiguated policy block added as an additional policy (\textit{Disambiguated}), (2) An explicit constraint block added to its prompt (\textit{Rule-Based}), and (3) the baseline setting with no policy disambiguation (\textit{No-ambiguated}) prompt. The other factors are held constant (i.e., Agreeable agent persona, GPT-4o mini customer simulator, and 20 max turns).

\begin{figure*}[t]
    \centering
    \includegraphics[width=0.8\textwidth]{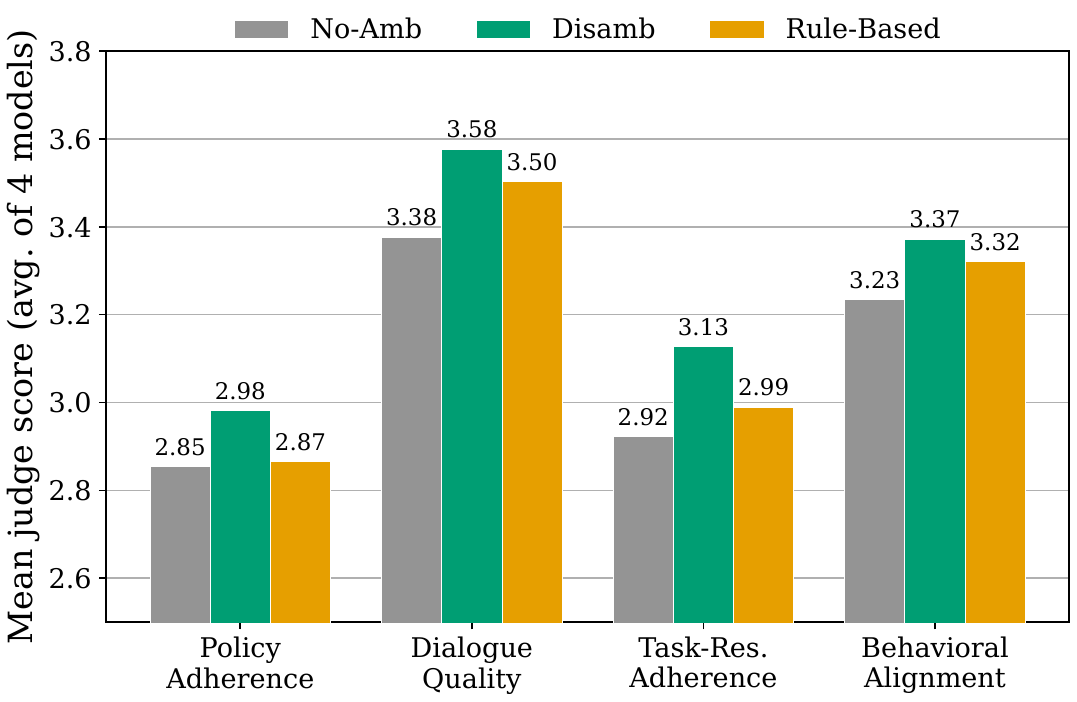} 
    \caption{Results of the controlled policy experiments (No-Amb: No Disambiguated Policies, Rule-Based: rule-based constraint block, and Disamb: Policy disambiguated) averaged across the models for different evaluation dimensions. The Y-axis represents the mean score across the evaluation dimensions in the X-axis. }
    \label{fig:ctrl_baseline}
\end{figure*}

The results, shown in Figure \ref{fig:ctrl_baseline}, represent the impact of policy ambiguities on the model's decision-making. Disambiguating the policy clearly improved every evaluation dimension on average (14 out of 16 model dimensions, with behavioral alignment of GPT 4.1 and Policy adherence of GPT-OSS as outliers). This establishes the impact of policy ambiguities, which negatively impact the decision quality of the models. Additionally, a rule-based control that hard-binds the resolution underperforms the disambiguated policy across all dimensions on average, because a fixed directive forces resolutions that contradict the model's own reasoning. Ambiguity, therefore, is a clear bottleneck on an agent's decision quality, and fixing it requires action on the policy level. Disambiguation improves every dimension, whereas a hard-rule directive helps compliant models but harms contextual reasoners, sometimes doing nothing at all.

\subsection{Run to run variance of the results}
\label{app:subsec:run_to_run}

To assess the reproducibility of the reported aggregate numbers under stochastic agent decoding, we re-scored K = 3 independent runs of each model on the 80-scenario stratified subsample with our judge pipeline. Table \ref{tab:run_to_run_stability} shows the aggregate metric means, which are highly stable across the runs. The largest standard deviation on any headline dimension is $\pm0.11$. In addition, bootstrapped 95\% confidence intervals over scenarios (1,000 resamples on the reported run) confirm that the per-model means are estimated precisely. The widest interval half-width across all metrics and models is $\pm0.09$, so the aggregate numbers are also insensitive to which scenarios are sampled.

\begin{table}[t]
\centering

\setlength{\tabcolsep}{3pt}
\resizebox{\columnwidth}{!}{%
\begin{tabular}{@{}lcccc@{}}
\toprule
\textbf{Metric}
& \textbf{GPT-5}
& \textbf{GPT-4.1}
& \textbf{Qwen}
& \textbf{GPT-OSS} \\
\midrule
Policy Adherence
& $2.94 \pm 0.03$
& $2.62 \pm 0.11$
& $2.16 \pm 0.03$
& $2.20 \pm 0.07$ \\

Dialogue Quality
& $3.77 \pm 0.01$
& $3.24 \pm 0.00$
& $2.47 \pm 0.09$
& $3.06 \pm 0.08$ \\

Behavioral Alignment
& $3.32 \pm 0.01$
& $3.14 \pm 0.05$
& $2.89 \pm 0.05$
& $3.18 \pm 0.03$ \\
\bottomrule
\end{tabular}%
}
\caption{Run-to-run aggregate stability on the 80-scenario subsample
($K=3$; mean $\pm$ SD).}
\label{tab:run_to_run_stability}
\end{table}


\subsection{Impact of policy ambiguity on shaping agent reasoning and resolution}
\label{app:subsec:policy_shaping}
\begin{figure*}[t]
    \centering
    \includegraphics[width=\textwidth]{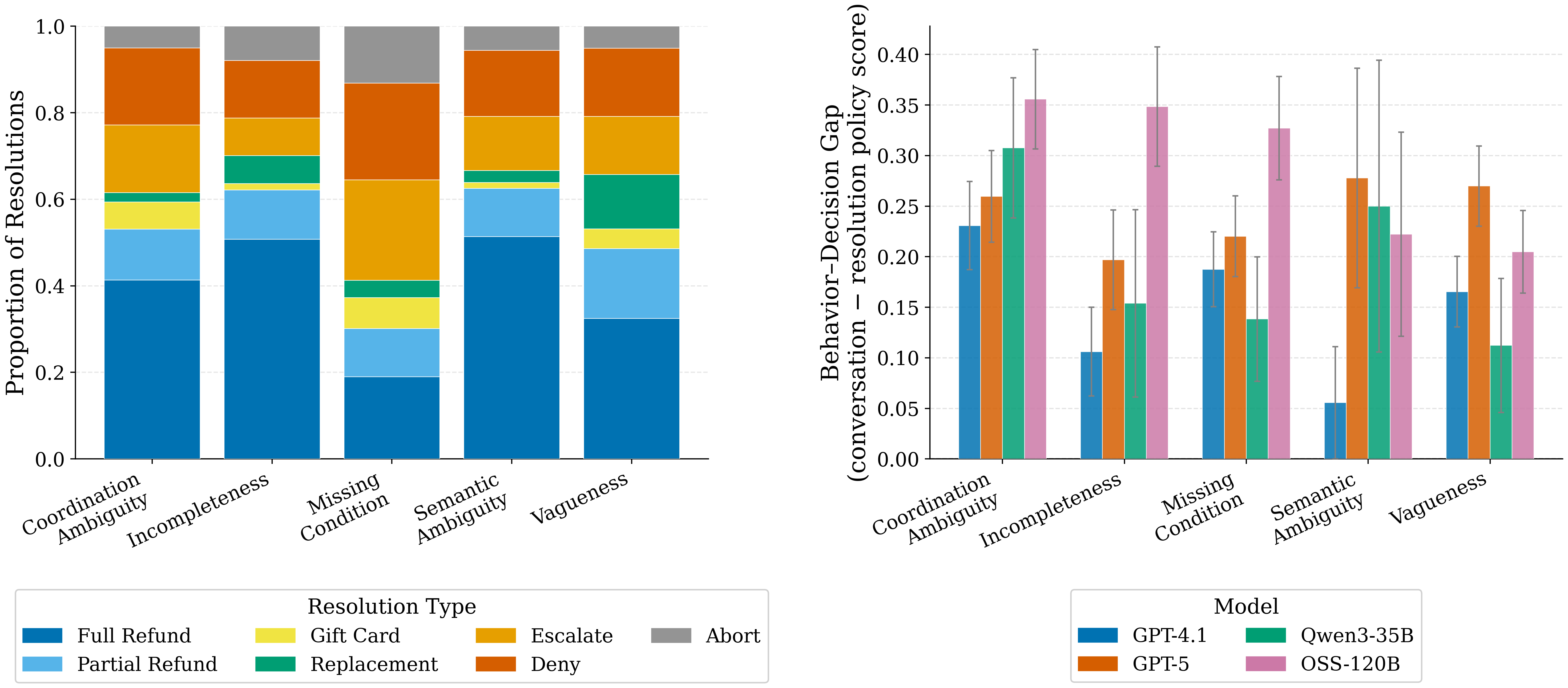} 
    \caption{
    Results of the ambiguity-type-to-resolution mapping. The left-hand side of the figure shows the ambiguity-type-to-resolution-proportion mapping, whereas the right-hand side plot shows the ambiguity-type-to-behavior-decision gap.
    }
    \label{fig:amb_to_res_mapping}
\end{figure*}

To examine how ambiguity type shapes resolution behavior, we analyze the proportion of outcomes per ambiguity type and the behavior-decision gap, defined as conversation policy adherence minus resolution policy adherence. Positive values indicate that the agent’s in-conversation reasoning is judged more policy-aligned than its final resolution. Results are shown in Figure~\ref{fig:amb_to_res_mapping}. 
Resolution distributions are strongly heterogeneous across ambiguity types, indicating that ambiguity structure systematically steers agents toward different strategies. The behavior-decision gap is positive across all ambiguity types and roughly consistent in magnitude, with the largest gap on Coordination Ambiguity (mean 0.29, 95\% CI [0.24, 0.34]; MWU $p = 0.002$), which is significantly higher than three of the four other types. This suggests that Coordination Ambiguity is particularly difficult for agents to resolve consistently with their stated reasoning. 
The well-populated ambiguity types (Coordination, Missing Condition, Vagueness, Incompleteness) are robustly estimated, carrying $\pm11 - 14pp$ confidence intervals. However, Semantic Ambiguity is underpowered with only 2 tasks, therefore, we report it with its wide interval ($ \pm17$ - 33pp).

\subsection{Intra-model stability analysis across runs}
\label{app:stability_analysis}

\begin{figure*}[t]
    \centering
    \includegraphics[width=\textwidth]{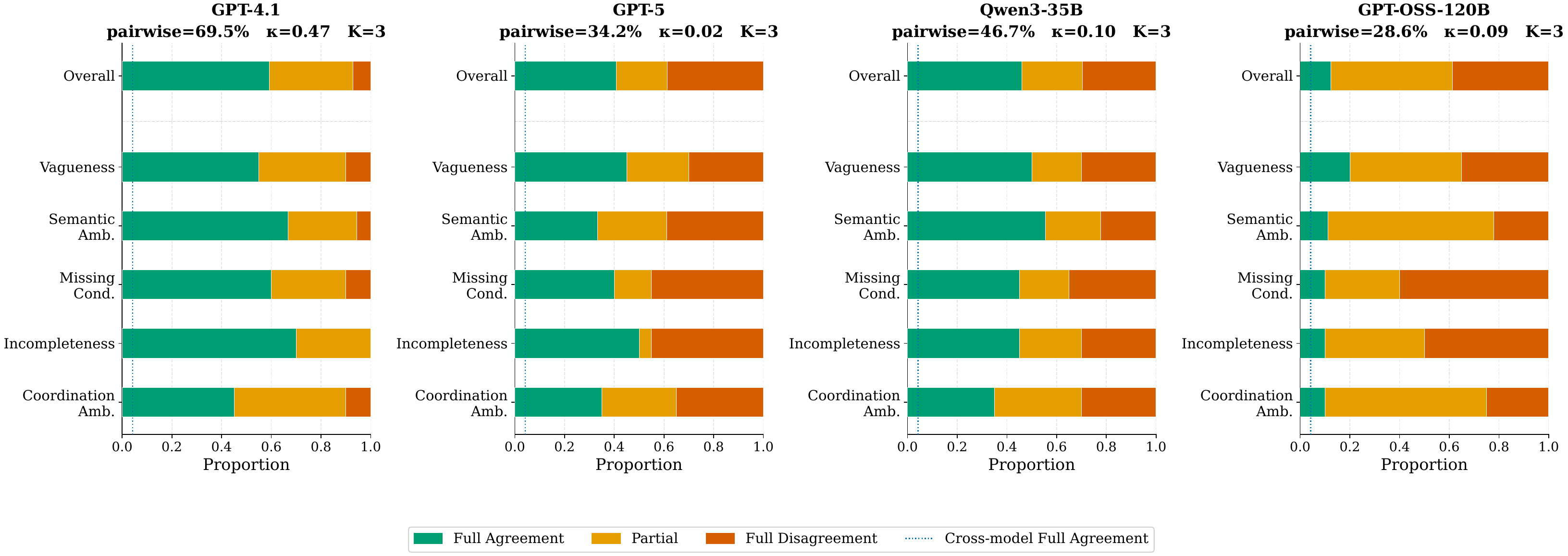} 
    \caption{
    Intra-model resolution stability across K independent runs on a stratified subsample. Each bar shows the proportion of scenarios reaching full agreement (all runs agree), partial agreement ($\geq 2$ runs agree), or full disagreement (all runs differ) for each ambiguity type. The dotted vertical line marks the cross-model full agreement rate}
    \label{fig:self_agreement_baseline}
\end{figure*}

To validate the cross-model agreement results, we ran each of the four models three times on a stratified subsample of 98 scenarios (20 per ambiguity type; 18 for Semantic Ambiguity), using the same agent persona and temperature as the original run. The findings are shown in Figure~\ref{fig:self_agreement_baseline}. Mean intra-model pairwise agreement is 44.7\%, ranging from 28.6\% (GPT-OSS-120B) to 69.5\% (GPT-4.1), with Semantic Ambiguity yielding the highest per-ambiguity-type agreement across models (51.1\%). This confirms that within-model variance alone produces low agreement rates (the low cross-model full agreement with 4.2\%), therefore reflects genuine resolution instability under ambiguous policy language, not solely behavioral differences between models. It is also important to note that the customer simulator (GPT-4o mini, temperature=0.8) is stochastic and contributes to the measured variance; isolating agent-only variance would require replaying fixed customer turns and is left for future work. The mean pairwise intra-model agreement for the models across the runs are shown in Table \ref{tab:intra_model_agreement}. Intra-model variance is high for GPT-5, Qwen3-35B, and GPT-OSS-120B. GPT-4.1 is the exception ($\kappa$=0.47, 7.1\% full disagreement). The three high-variance models show that under policy ambiguity, re-running the same model is nearly as unpredictable as consulting a different model. 

\begin{table}[t]
\centering
\small
\setlength{\tabcolsep}{3pt}
\begin{tabular}{@{}lcccc@{}}
\toprule
\textbf{Ambiguity}    & \textbf{GPT-4.1} & \textbf{GPT-5} & \textbf{Qwen} & \textbf{OSS} \\
\midrule
Coordination          & 60.0 & 35.3 & 46.3 & 31.7 \\
Incompleteness        & 80.0 & 25.6 & 47.9 & 23.3 \\
Missing Cond.         & 64.7 & 31.2 & 30.3 & 20.0 \\
Semantic              & 75.9 & 42.6 & 52.4 & 33.3 \\
Vagueness             & 66.7 & 33.3 & 52.1 & 35.0 \\
\midrule
\textbf{Overall}      & \textbf{69.5} & \textbf{34.2} & \textbf{46.7} & \textbf{28.6} \\
\bottomrule
\end{tabular}
\caption{Intra-model pairwise self-agreement (\%) across three independent runs (K=3, $n{=}98$). Qwen = Qwen3-35B-A3B; OSS = GPT-OSS-120B.}
\label{tab:intra_model_agreement}
\end{table}

A natural concern is whether the low cross-model agreement (4.2\% full agreement) reflects ambiguity-driven reasoning differences or merely shared training conventions among same-lab models, since three of our four evaluated models originate from OpenAI. We test this directly by comparing intra-OpenAI to cross-lab pairwise agreement on the 98-scenario stratified subsample. The two OpenAI models, GPT-4.1 and GPT-5, agree on only 21.1\% of scenarios, which is below the cross-lab mean of 26.7\%, and lower than GPT-4.1's agreement with Qwen3-35B (41.7\%). Notably, intra-OpenAI cross-model agreement (21.1\%) is also lower than either model's intra-model self-agreement across three runs (69.5\% for GPT-4.1, 34.2\% for GPT-5), meaning that switching between same-lab models disrupts agreement more than reseeding a single model. Shared lab origin therefore provides no convergence benefit, indicating that resolution disagreement is driven by policy ambiguity rather than cross-lab behavioral divergence.

\subsection{Cross-judge Agreement}
\label{app:cross-judge-agreement}

To test the judge robustness and justify selection, we test with two more judge models (Claude Sonnet 4.6 and Mistral-Medium-3.5-128B) on a subsample (80 scenarios × 4 models, stratified by ambiguity type). This test verifies that our findings are not an artifact of a single judge family. We re-score a fixed set of 80 held-out transcripts, spanning four agent models (GPT-4.1, GPT-5, GPT-OSS, and Qwen; 320 items total), and all judges used the same rubric and scored each transcript on four dimensions (Policy Adherence, Dialogue Quality, Behavioral Alignment, and Interest Alignment). Because no human gold labels exist for these transcripts, we assess reliability through inter-judge agreement. For each dimension we report pairwise Spearman $\rho$ and quadratic-weighted Cohen's $\kappa$ on all commonly rated items. Table \ref{tab:cross_judge_agreement} summarizes the results. GPT-5.4 and Claude agree closely (e.g., Dialogue Quality Spearman = 0.80), whereas Mistral is a systematic outlier, as it preserves rank order (Spearman ≈ 0.51 on Policy Adherence) but assigns markedly more lenient absolute scores (mean 4.63 vs. ≈2.6), collapsing weighted $\kappa$ to ≈0.08. We find that independent judges reproduce the same quality ordering. Furthermore, the judges differ in calibration (assign the same absolute number on the 1-5 scale): Claude is the harshest (3.10/5), Mistral is most lenient (4.17/5), and GPT-5.4 is the median (3.20/5) while keeping near-top agreement. As for the self-provider bias, we find no pooled evidence of GPT-5.4 favoring OpenAI agents in this difference-in-differences analysis. Therefore based on the balanced performance and no evidence of self-provider bias, we choose GPT-5.4 as our preferred judge.

\begin{table}[t]
\centering
\setlength{\tabcolsep}{3.5pt}
\small
\begin{tabular}{lccc}
\toprule
\textbf{Dimension}
& \textbf{G5.4--C}
& \textbf{G5.4--M}
& \textbf{C--M} \\
\midrule
Policy Adherence     & 0.65 / 0.54 & 0.51 / 0.08 & 0.51 / 0.09 \\
Dialogue Quality     & 0.80 / 0.76 & 0.71 / 0.29 & 0.72 / 0.34 \\
Behavioral Alignment & 0.75 / 0.49 & 0.59 / 0.25 & 0.58 / 0.35 \\
Interest Alignment   & 0.42 / 0.39 & 0.35 / 0.31 & 0.63 / 0.63 \\
\bottomrule
\end{tabular}
\caption{Cross-judge agreement (Spearman $\rho$ / quadratic-weighted Cohen's $\kappa$), 80 scenarios $\times$ 4 models. \textit{Note:} G5.4 = GPT-5.4, C = Claude, M = Mistral.}
\label{tab:cross_judge_agreement}
\end{table}

\section{Ambiguity Definition}
\label{app:ambiguity_types}
As mentioned in Section \ref{sec:benchmark_construction}, we follow the works of \citet{massey2014identifying} and \citet{gervasi2019ambiguity}. A single policy clause can exhibit several ambiguity types at once. Our tasks are constructed with multiple types rather than one per task (mean 2.95, median 3). The ambiguity types are defined as follows.

\begin{itemize}
    \item \textbf{Vagueness}: Occurs when a term, statement, or clause admits borderline cases or relative interpretation. 
    \item \textbf{Incompleteness:} Occurs when a statement or clause fails to provide enough information to have a single clear interpretation, leaving gaps in specification. 
    \item \textbf{Coordination Ambiguity:} Ambiguity caused by coordinating conjunctions, especially when multiple conjunctions appear in one clause; modifiers can attach to one or multiple coordinated elements; or lists allow multiple logical interpretations.
    \item \textbf{Missing Condition:} Ambiguity where an ``if" or implicit condition is not followed by a clearly defined ``then, ``else," or ``otherwise" clause, or where outcomes, exceptions, deadlines, responsibilities, or consequences are omitted for specific scenarios.
    \item \textbf{Semantic Ambiguity} Occurs when a sentence or clause has more than one interpretation based entirely on surrounding context, even when grammar and individual word meanings are clear.
\end{itemize}

\begin{table}[h]
\centering
\scriptsize
\setlength{\tabcolsep}{4pt}
\renewcommand{\arraystretch}{1.15}
\begin{tabularx}{\linewidth}{@{}l X X@{}}
\toprule
\textbf{Ambiguity Type} & \textbf{Policy Clause} & \textbf{Ambiguity Description} \\
\midrule
\textbf{Vagueness}
& \textit{``Most products may be returned for a refund, replacement, or exchange within 30 days of delivery, provided they remain in original or unused condition.''}
& The policy does not define what qualifies as ``original or unused condition.'' \\
\addlinespace
\textbf{Semantic Ambiguity}
& \textit{``You may be charged a fee if you do not drop off or complete a carrier pickup by the `return by date'.''}
& Late-fee logic depends on the operational definition of ``return by date'' and what constitutes ``drop off'' vs.\ ``complete a carrier pickup.'' \\
\addlinespace
\textbf{Incompleteness}
& \textit{``Once an item has been returned, it cannot be sent back to the customer, except for certain Luxury Stores items that may incur 100\% damage fees.''}
& The policy does not identify which Luxury Stores items or circumstances fall under this exception, nor does it explain the applicable process. \\
\addlinespace
\textbf{Coordination Ambiguity}
& \textit{``Returned items should be in original or unused condition, with tags attached and hygiene seals and liners intact, and in the original manufacturer's packaging.''}
& There is ambiguity regarding whether failure to comply with any sub-requirement disqualifies the return or merely reduces the refund. \\
\addlinespace
\textbf{Missing Condition}
& \textit{``Items from different orders should not be included in the same return.''}
& The clause does not state the consequence of doing so, such as processing delays, rejection, risk of loss, or fees, nor does it explain how to correct the issue after it occurs. \\
\bottomrule
\end{tabularx}
\caption{Examples of ambiguity types, corresponding policy clauses, and their ambiguity descriptions. Policy clauses are extracted and paraphrased from Amazon's return policy.}
\label{tab:example_policy_ambiguities}
\end{table}

Table \ref{tab:example_policy_ambiguities} shows examples of policy ambiguity types along with the clause and the ambiguity description. The ambiguity types are directly connected to corresponding policy clauses in the table. The policy clauses are refined from Amazon return policy to use as examples here, and the ambiguity descriptions are LLM generated and human validated. We discuss the process of collecting policy ambiguities in Section \ref{subsec:dataset_construction}.

\section{Evaluation Metrics}
\label{app:eval_dimensions}
\begin{figure*}[t]
    \centering
    \includegraphics[width=\textwidth]{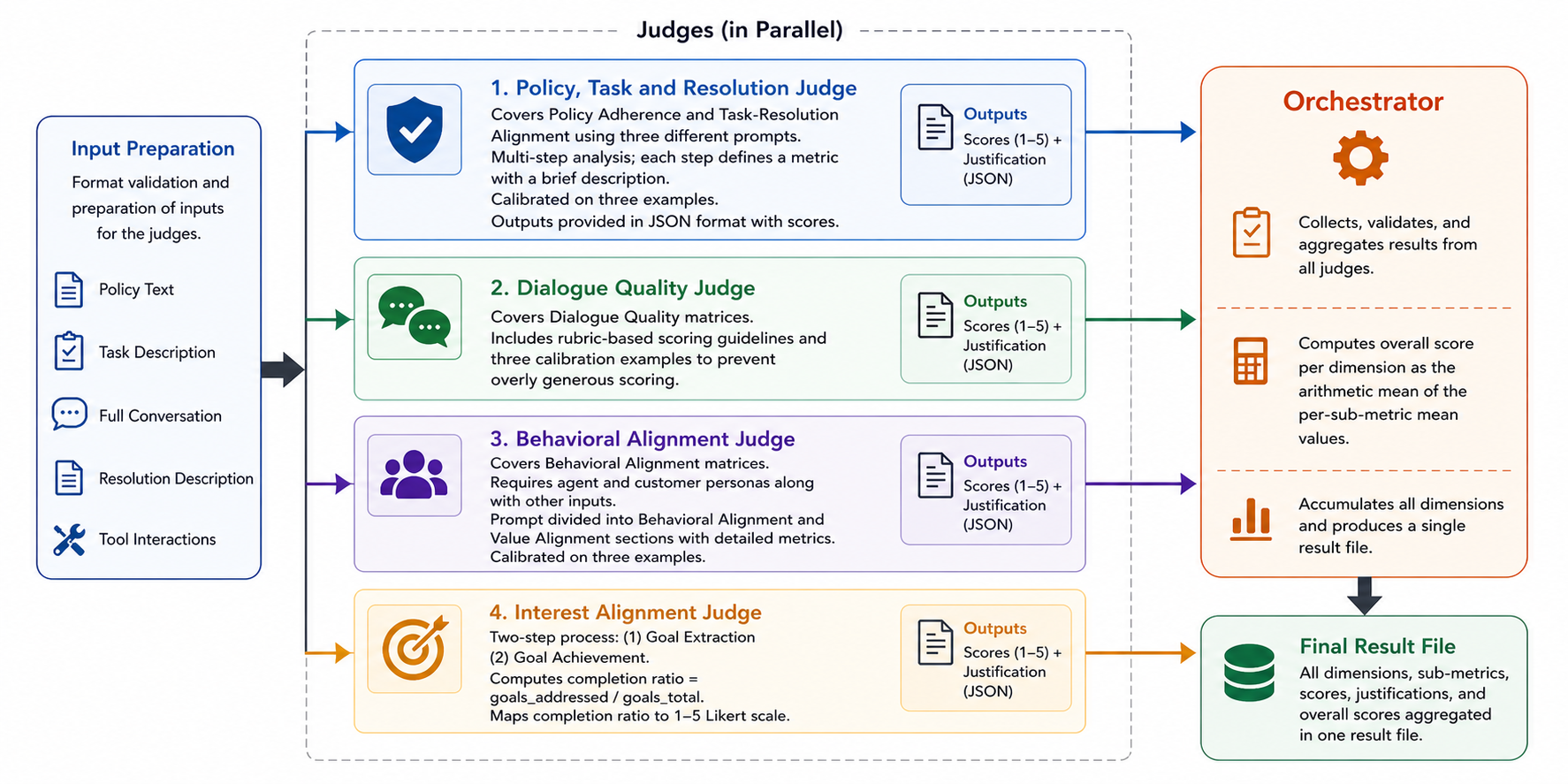}
    \caption{
    Overview of the evaluation pipeline. Each of the judges (implemented in parallel) represents the evaluation dimensions.
    }
    \label{fig:eval_pipeline}
\end{figure*}

In this section, we define all evaluation metrics and their definitions. We present the metrics by dimensions, as defined in Section \ref{subsec:eval_pipeline} and the judges shown in Figure \ref{fig:eval_pipeline}.

\paragraph{Policy Adherence}
\begin{itemize}
    \item \textbf{Policy Support (Conversation)} [Scale: 1--5; Type: conversation]: This metric evaluates whether the agent’s responses align with the provided return policy and related policies across the conversation.

    \item \textbf{Policy Support (Per-Turn )} [Scale: 1--5; Type: turn-by-turn]: This metric evaluates whether any policy claim made in this turn is justified per the provided policy text.

    \item \textbf{Policy Support (Resolution)} [Scale: 1--5; Type: resolution]: This metric evaluates whether the outcome is clearly justified by the return policy based on the provided facts.

    \item \textbf{Reasoning Quality} [Scale: 1--5; Type: resolution]: This metric evaluates whether the outcome explanation demonstrates appropriate reasoning given the task complexity and policy issues.
\end{itemize}

\paragraph{Task-resolution Adherence}
\begin{itemize}
    \item \textbf{Outcome Addresses Customer Reason} [Scale: 1--5; Type: resolution]: This metric evaluates whether the outcome directly addresses the customer’s stated reason for the return.

    \item \textbf{Consistent with Task Constraints} [Scale: 1--5; Type: resolution]: This metric evaluates whether the resolution is consistent with the stated conditions such as usage, timing, and bundle details.
\end{itemize}

\paragraph{Dialogue/Conversation Quality}
\begin{itemize}
    \item \textbf{Conversation Consistency} [Scale: 1--5; Type: conversation]: This metric evaluates whether the agent’s responses are logically consistent with the user’s stated intent, previously shared information, and overall conversation context.

    \item \textbf{Resolution-Oriented Responses} [Scale: 1--5; Type: conversation]: This metric evaluates whether the conversation progresses in a structured manner toward resolving the user’s return request.

    \item \textbf{Conversation to Resolution Mapping} [Scale: 1--5 per Res.; Type: resolution]: This metric evaluates how effectively the conversation leads to the final resolution, ensuring each step logically contributes to the stated result.

    \item \textbf{Verbosity} [Scale: 1--5 per Conv; Type: conversation]: This metric evaluates whether the length and level of detail in the agent’s responses are appropriate for addressing the user’s needs. It also evaluates whether response length and detail are appropriate given the customer’s preceding message.

    \item \textbf{Per-Turn Contradiction Score (Turn-Level Consistency)} [Scale: 1--5 per turn; Type: turn-by-turn conversation]: This metric evaluates whether the agent’s statement in this turn contradicts anything stated in any prior turn.

    \item \textbf{Consistency Drift} [Scale: 1--5 per turn; Type: turn-by-turn conversation]: This metric evaluates whether the conversation as a whole has accumulated contradictions up to and including this turn, relative to the prior conversation state.

    \item \textbf{Arguments Match Conversation/Execution Fidelity} [Scale: 1--5 per Conv; Type: conversation]: This metric evaluates whether the tool call arguments collectively reflect what was agreed upon in the conversation.

    \item \textbf{Resolution Description Fidelity [Say-Do Gap (Outcome+conversation)]} [Scale: 1--5 per Res./Conv; Type: resolution+conversation]: This metric evaluates whether the stated resolution description accurately reflects the tool call arguments executed. It also evaluates whether each commitment made during the conversation is reflected in a corresponding backend tool call action.
\end{itemize}

\paragraph{Interest Alignment}
\begin{itemize}
    \item \textbf{Interest Alignment Customer} [Scale: 1--5; Type: conversation]: Computed by LLM-extracting per-task customer goals and judging satisfaction independently for each, then taking the proportion satisfied.

    \item \textbf{Interest Alignment Company} [Scale: 1--5; Type: conversation]: This metric evaluates the degree to which the agent’s behavior and outcome favor the company’s interest.
\end{itemize}

\paragraph{Behavioral Alignment}
\begin{itemize}
    \item \textbf{Agent Tone Consistency} [Scale: 1--5; Type: conversation]: This metric evaluates whether the agent’s language and demeanor match the assigned agent persona throughout the conversation.

    \item \textbf{Agent Behavioral Alignment} [Scale: 1--5; Type: conversation]: This metric evaluates whether the agent’s decisions such as leniency, strictness, and escalation tendency reflect the assigned persona’s expected disposition.

    \item \textbf{Customer Tone Consistency} [Scale: 1--5; Type: conversation]: This metric evaluates whether the simulated customer’s language and communication style match the assigned persona traits.

    \item \textbf{Customer Behavioral Alignment} [Scale: 1--5; Type: conversation]: This metric evaluates whether the customer’s actions such as persistence, acceptance, and escalation reflect the assigned persona description.

    \item \textbf{Interest Alignment Customer} [Scale: 1--5; Type: conversation]: This metric evaluates the degree to which the agent’s behavior and outcome favor the customer’s interest.

    \item \textbf{Interest Alignment Company} [Scale: 1--5; Type: conversation]: This metric evaluates the degree to which the agent’s behavior and outcome favor the company’s interest.

    \item \textbf{Conformity [Value Rating (Schwartz)]} [Scale: 1--5; Type: conversation]: This metric evaluates restraint of actions, inclinations, and impulses likely to upset or harm others and violate social expectations or norms.

    \item \textbf{Benevolence [Value Rating (Schwartz)]} [Scale: 1--5; Type: conversation]: This metric evaluates preserving and enhancing the welfare of those with whom one is in frequent personal contact, the `in-group'. Benevolence values emphasize voluntary concern for others’ welfare, including being helpful, honest, forgiving, responsible, loyal, true friendship, and mature love, as well as sense of belonging, meaning in life, and a spiritual life.

    \item \textbf{Self-Direction [Value Rating (Schwartz)]} [Scale: 1--5; Type: conversation]: This metric evaluates the degree to which the agent exercises independent judgment in ambiguous or complex situations rather than mechanically applying rules.

    \item \textbf{Security [Value Rating (Schwartz)]} [Scale: 1--5; Type: conversation]: This metric evaluates the degree to which the agent prioritizes safety, stability, and risk avoidance for the company. It also evaluates safety, harmony, and stability of society, of relationships, and of self.

    \item \textbf{Universalism [Value Rating (Schwartz)]} [Scale: 1--5; Type: conversation]: This metric evaluates understanding, appreciation, tolerance, and protection for the welfare of all people and for nature.
\end{itemize}

\paragraph{Evaluation of Model performance across different dimensions}

\begin{table*}[ht]
\centering
\footnotesize
\setlength{\tabcolsep}{4pt}
\renewcommand{\arraystretch}{1.15}
\begin{tabularx}{\linewidth}{Y R{0.10\linewidth} R{0.10\linewidth} R{0.10\linewidth} R{0.10\linewidth}}
\toprule
\textbf{Metric and definition} &
\multicolumn{1}{l}{\textbf{GPT-4.1}} &
\multicolumn{1}{l}{\textbf{GPT-5}} &
\multicolumn{1}{l}{\textbf{Qwen-3.6}} &
\multicolumn{1}{r}{\textbf{GPT-OSS}} \\
\midrule

\multicolumn{5}{l}{\textbf{\textit{Policy Adherence}}} \\
\addlinespace[2pt]
\textbf{Policy Support (Conversation)}: Agent responses align with return policy &
3.98 & \textbf{4.02} & 3.10 & 3.73 \\

\textbf{Policy Support (Resolution)}: Outcome justified by policy &
\textbf{3.81} & 3.78 & 2.70 & 3.43 \\

\textbf{Policy Support (Per-Turn)}: Per-turn justification of policy claims &
4.65 & \textbf{4.83} & 3.55 & 4.68 \\

\midrule

\multicolumn{5}{l}{\textbf{\textit{Dialogue Quality}}} \\
\addlinespace[2pt]
\textbf{Verbosity}: Appropriate response length and detail &
3.85 & \textbf{3.90} & 3.22 & 3.67 \\

\textbf{Consistency Drift (Agent)}: Accumulated contradictions across conversation state &
4.95 & \textbf{4.98} & 4.68 & 4.81 \\

\textbf{Resolution-Oriented Responses}: Progresses toward resolving return request &
\textbf{3.99} & 3.96 & 3.39 & 3.80 \\
\midrule

\multicolumn{5}{l}{\textbf{\textit{Behavioral Alignment}}} \\
\addlinespace[2pt]
\textbf{Agent Behavioral Alignment}: Decisions reflect assigned agent persona &
3.86 & \textbf{4.12} & 3.50 & 3.83 \\

\textbf{Customer Behavioral Alignment}: Actions reflect customer persona description &
4.52 & \textbf{4.85} & 4.48 & 4.75 \\

\textbf{Top Schwartz Value}: Highest rated Schwartz value &
Universalism & Conformity & Universalism & Universalism \\

\midrule
\multicolumn{5}{l}{\textbf{\textit{Interest Alignment}}} \\
\addlinespace[2pt]
\textbf{Company Interest Alignment}: Outcome favors company interest &
3.19 & \textbf{3.51} & 2.56 & 3.34 \\

\textbf{Customer Goal Alignment}: Fraction of customer goals addressed &
3.32 & \textbf{3.41} & 3.14 & 2.79 \\
\midrule

\multicolumn{5}{l}{\textbf{\textit{Task-Resolution Adherence}}} \\
\addlinespace[2pt]
\textbf{Outcome Addresses Customer Reason}: Directly addresses stated return reason &
4.23 & \textbf{4.32} & 3.14 & 3.62 \\

\bottomrule
\end{tabularx}

\vspace{0.6em}
\caption{Evaluation of model performance across policy adherence, dialogue quality, behavioral alignment, and task-resolution adherence metrics.}
\label{tab:main_results}
\end{table*}

Table \ref{tab:main_results} presents the results of the metrics defined in our benchmark for the models that we test. A part of the table is represented in Figure~\ref{fig:all_performance_spider}, where we can draw direct comparisons among metrices of policy adherence, dialogue quality, behavioral alignment, and task-resolution adherence dimensions.

\section{Task diversity and Stratification}
\label{app:task_diversity}
To ensure task diversity and complexity, we use a hybrid pipeline that combines LLM-as-Judge and human selection to stratify and select tasks according to the following criteria. The pipeline is designed so that the retained tasks are spread across distinct situations, policy clauses, product domains, and condition types.


\begin{itemize}
    \item \textbf{Ambiguities:} The number of ambiguities contained in the task. The Ambiguities criterion requires more than one type per task, so multiple ambiguity types per task are the norm by construction.
    
    \item \textbf{Complexity:} Complexity is measured using direct ambiguity patterns \citep{niwa2024ambignlg} (Context, Keywords, Planning, and Theme), and indirect patterns \citep{mannekote2025making} (Justification-embedded requests, Elaboration-obscured intent, and Implicit slot values).
    
    \item \textbf{Diversity:} Task diversity is defined across four dimensions: scenario diversity (how unique the situational setup is), policy coverage breadth (how many distinct policy clauses are invoked), product domain variety (how diverse the product types involved are), and condition ambiguity depth (how interpretively complex the item conditions are).
\end{itemize}

We collect an initial pool of 500 tasks using the task collection mechanism described in Section~\ref{subsec:dataset_construction}. These tasks are processed through the pipeline below and, after stratification, filtered by an expert to 60 tasks. Since this step is an analysis step, it is not included in the main dataset construction pipeline. The stratification pipeline proceeds as follows.

\begin{enumerate}
    \item \textbf{Data Preparation:} We load and prepare the policy ambiguities and tasks, compute sentence embeddings for the tasks, parse the policy issues column, and build an ambiguity lookup dictionary.
    
    \item \textbf{Ambiguity Type Classification:} We implement a two-stage classification process. The first stage performs exact matching, checking whether the issue string exactly matches or is a substring of any key in the lookup dictionary. The second stage applies a three-step re-ranker: (1) cosine similarity between the issue string and dictionary keys to select the top 20 candidates, (2) cross-encoder reranking to score each (issue, candidate) pair and select the highest-scoring match, and (3) an unmatched tier for instances that do not satisfy either stage. This forms the primary type of ambiguity for each task. In practice, all 500 tasks were classified via the cross-encoder stage, with an exact match rate of 0\% and an unmatched rate of 0\%, confirming full coverage.

    \item \textbf{LLM-Judged Complexity and Diversity:} We instantiate an LLM-as-a-Judge using the complexity and diversity definitions above, with a 1--5 scoring scale. Three few-shot anchor examples (low, medium, and high) are provided to condition the judge.
    
    \item \textbf{Stratification:} The four diversity sub-dimension Scores are aggregated into a diversity composite by taking their mean, which is then mapped to a three-tier system: Low (1.00--2.33), Medium (2.34--3.66), and High (3.67--5.00). The stratum label is defined as $\textit{primary\_ambiguity\_type}\textit{\_}\textit{diversity\_tier}$, yielding 9 strata, of which 8 are well-populated ($n \geq 5$); only \textsc{Semantic Ambiguity\_High} is flagged as thin ($n = 2$). The Spearman $\rho$ between complexity and diversity composite is 0.743, indicating moderate but acceptable construct separation.
\end{enumerate}

Following stratification, an expert selects tasks from each stratum, yielding 60 tasks in total. To confirm diversity, we run a 3-run resolution-collection setup under identical settings and model, prompting the model to provide a resolution as a customer service agent would. We find that 35 of the 60 tasks produce at least one differing output across runs. Based on this, we retain 5 outcome-invariant tasks (in addition to the 35) to complete the final selection of 40 benchmark tasks, while preserving diversity and complexity across products and ambiguity types.

\section{Persona Description}
\label{app:persona}
In this section, we describe the characteristics we adopt for the user and agent personas.
\subsection{User persona}
\label{app:user persona}
As mentioned in Section \ref{subsec:dataset_construction}, the user persona is augmented from the Persona-Hub dataset \citep{ge2024scaling}, and contains the following information.

\begin{itemize}
    \item \textbf{Name:} Name of the persona.
    
    \item \textbf{Age-range:} Age range of the persona, ideally placed within a range of 18-65, with 5-year intervals.
    
    \item \textbf{Location:} Country where the persona resides.
    
    \item \textbf{Gender:} Gender of the persona, should be consistent with the name provided, and can be Male, Female, Non-binary, etc.
    
    \item \textbf{Job Sector:} The industry or field in which the persona works, e.g., Technology, Healthcare, Education, Finance, etc.
    
    \item \textbf{Income range:} The estimated annual income range of the persona, e.g., 30,000-50,000, 50,000-70,000, etc., with currency appropriate to the location.
    
    \item \textbf{Person description:} A brief description of the persona, including their interests, lifestyle, and any other relevant details that make them unique.
    
    \item \textbf{Language Style:} Choose one of the following:
    \begin{itemize}
        \item \textbf{Frozen:} Very formal and structured language, often used in official documents.
        \item \textbf{Formal:} Professional and polite language, suitable for business communications.
        \item \textbf{Consultative:} Balanced and neutral language, appropriate for general conversations.
        \item \textbf{Casual:} Informal and friendly language, often used in everyday interactions.
        \item \textbf{Intimate:} Very casual and personal language, used among close friends and family.
    \end{itemize}
    
    \item \textbf{Personality Style:} Likert scale for Big Five Personality Traits: Openness, Conscientiousness, Extraversion, Agreeableness, and Neuroticism.
    \begin{itemize}
        \item \textbf{Openness (1-5):} The degree to which the persona is open to new experiences and ideas.
        \item \textbf{Conscientiousness (1-5):} The level of organization, dependability, and discipline the persona exhibits.
        \item \textbf{Extraversion (1-5):} The extent to which the persona is outgoing, sociable, and energetic.
        \item \textbf{Agreeableness (1-5):} The degree of kindness, empathy, and cooperation the persona shows towards others.
        \item \textbf{Neuroticism (1-5):} The level of emotional stability and tendency towards anxiety or moodiness the persona has.
    \end{itemize}
    
    \item \textbf{Communication style:} Choose one of the following:
    \begin{itemize}
        \item \textbf{Direct:} Clearly and explicitly stating one’s feelings, needs and wants; the speaker says what they mean.
        \item \textbf{Indirect:} Using verbal messages that imply but does not clearly say speakers’ true intentions in terms of their wants, needs, and goals in the discourse situation.
        \item \textbf{Linear:} Structured communication focusing on one topic at a time, with clear objectives, and limited interaction.
        \item \textbf{Non-Linear:} Dynamic communication that can involve multiple focuses and is interactive.
        \item \textbf{Being centered:} Focuses more on people, the quality of the social interactions, emotional connection, and interpersonal relationship.
        \item \textbf{Doing centered:} Focuses more on the work, results, efficiency, and achieving goals.
        \item \textbf{Literate:} Relies on written and relatively formal language through emails, letters, charts and reports.
        \item \textbf{Oral:} Involves spoken language through conversations, meetings, presentations, and phone calls.
    \end{itemize}
    
    \item \textbf{Persona\_id:} A unique identifier for the persona. It must follow this exact format: \texttt{Firstname\_JobSector\_CountryCode\_NN}, e.g., \texttt{John\_Legal\_USA\_01}, \texttt{Emily\_Healthcare\_CAN\_02}. It should be short, with a maximum of 30 characters total. Use standard 2--3 letter country codes, e.g., USA, UK, CAN, CHN, etc.
    
    \item \textbf{Purchase\_history:} A brief overview of the persona's typical purchasing behavior, including preferred product categories, shopping frequency, and spending habits. It should contain the number of orders and returns in the last year.
\end{itemize}

Below is an example of User persona. Note that the persona information is summarized in the example below to get a better overview.

\lstdefinestyle{jsonbox}{
    basicstyle=\ttfamily\scriptsize,
    breaklines=true,
    breakatwhitespace=true,
    columns=fullflexible,
    keepspaces=true,
    showstringspaces=false,
    frame=none
}

\begin{tcolorbox}[
    colback=gray!8,
    colframe=gray!55,
    boxrule=0.5pt,
    arc=2mm,
    left=6pt,
    right=6pt,
    top=6pt,
    bottom=6pt,
    width=\linewidth
]
\begin{lstlisting}[style=jsonbox]
{
  "Name": "Latasha Edwards",
  "Age-range": "35-40",
  "Location": "United States",
  "Gender": "Female",
  "Job Sector": "Public Policy and Social Research",
  "Income range": "85,000-110,000 USD",
  "Person description": "Policy analyst focused on poverty reduction, child advocacy, racial and ethnic disparities, and systemic inequality.",
  "Language Style": "Consultative",
  "Personality Style": {
    "Openness": 5,
    "Conscientiousness": 4,
    "Extraversion": 3,
    "Agreeableness": 5,
    "Neuroticism": 3
  },
  "Communication style": "Being centered",
  "Persona_id": "Latasha_PublicPolicy_USA_01",
  "Purchase_history": {
    "Orders_last_year": 25,
    "Returns_last_year": "Not specified",
    "Overview": "Typical purchasing categories, shopping frequency, and spending habits are not specified."
  }
}
\end{lstlisting}
\end{tcolorbox}

\subsection{Agent persona}
\label{app:agent_personas}
We create the agent persona with prioritization to the personality,communication, decision approach and behavioral description. Using Big-5 personality style \citep{big5personality}, we introduce 5 agent personas ranging from Direct to Very helpful, as described below.

\begin{itemize}
    \item \textbf{DIRECT:} An agent persona characterized by low openness, high conscientiousness, low extraversion, low agreeableness, and low neuroticism. This agent communicates in a straightforward, concise, and to-the-point manner. Decisions are made strictly according to policy, without unnecessary elaboration. The agent states facts clearly, avoids sugarcoating outcomes, and focuses on efficient resolution.

    \item \textbf{FAIR:} An agent persona characterized by moderate openness, high conscientiousness, moderate extraversion, moderate agreeableness, and low neuroticism. This agent communicates in a balanced, objective, and procedural manner. Decisions are made by treating all cases equally and following established procedures consistently. The agent explains reasoning transparently, applies policy uniformly, and acknowledges both sides. 

    \item \textbf{AGREEABLE:} An agent persona characterized by moderate openness, moderate conscientiousness, moderate extraversion, high agreeableness, and low neuroticism. This agent communicates warmly and empathetically while seeking common ground. Decisions aim for win--win solutions within policy boundaries. The agent validates customer feelings, offers alternatives, and emphasizes positive outcomes. 

    \item \textbf{HELPFUL:} An agent persona characterized by high openness, high conscientiousness, moderate extraversion, high agreeableness, and low neuroticism. This agent communicates proactively, thoroughly, and with a solution-oriented style. Decisions involve going the extra mile within policy and anticipating customer needs. The agent offers additional assistance, explains all options fully, and follows up on details.

    \item \textbf{VERY\_HELPFUL:} An agent persona characterized by high openness, low-to-moderate conscientiousness, high extraversion, very high agreeableness, and low neuroticism. This agent communicates enthusiastically, accommodatively, and with a customer-first orientation. Decisions prioritize customer satisfaction and may involve bending small rules or making exceptions. The agent seeks workarounds, advocates for the customer, and may favorably approve edge cases.
\end{itemize}


\section{Conversation and Resolution Structure}
\label{app:res_structure}
As mentioned in Section \ref{sec:benchmark_construction}, we follow the conversation and resolution structure in full-duplex conversations with tool interactions, in which both the user and the customer service agent can interact with tools. We provide examples of the conversation in the Supplementary material.

\paragraph{Conversation Structure}
The conversation array is a sequence of turns with a turn discriminator. The turn types are: dialogue turns (conversation statements from the Customer and the agent), Tool call turns (one or more tool invocations), and tool result turns (the environment's response to the preceding tool call). The pattern for a tool interaction is always tool call --> tool result --> dialogue turn. The JSON format of the turns are shown below.

\lstdefinestyle{jsonbox}{
    basicstyle=\ttfamily\scriptsize,
    breaklines=true,
    breakatwhitespace=true,
    columns=fullflexible,
    keepspaces=true,
    showstringspaces=false,
    frame=none
}

\begin{tcolorbox}[
    colback=gray!8,
    colframe=gray!55,
    boxrule=0.5pt,
    arc=2mm,
    left=6pt,
    right=6pt,
    top=6pt,
    bottom=6pt,
    width=\linewidth
]
\begin{lstlisting}[style=jsonbox]
{
  "Customer_or_Agent_turn": {
    "turn": "customer",
    "message": "Hello, I would like to return some items from my recent order...",
    "tool_calls": null,
    "tool_result": null
  },

  "Tool_call_turn": {
    "turn": "tool_call",
    "message": null,
    "tool_calls": [
      {
        "tool_name": "get_order_details",
        "tool_call_id": "call_002",
        "arguments": {
          "order_id": "ORD-123456",
          "customer_id": "cust_alastair_macgregor",
          "include_tracking": false,
          "detail_type": "full"
        }
      }
    ],
    "tool_result": null
  },

  "Tool_result_turn": {
    "turn": "tool_result",
    "message": null,
    "tool_calls": null,
    "tool_result": {
      "tool_call_id": "call_002",
      "tool_name": "get_order_details",
      "result": {
        "order_id": "ORD-123456",
        "items": ["..."],
        "return_window": "30 days from delivery"
      }
    }
  }
}
\end{lstlisting}
\end{tcolorbox}

\paragraph{Resolution Structure}
The resolution is a JSON object with two fields: Resolution type and Resolution description. Resolution type shows the resolution obtained by the agent (one of the seven values: Full refund, Partial refund, Deny refund, etc., mentioned in Section \ref{subsec:problem_setup}). The resolution description is a free-text narrative written by the agent to justify the resolution. It contains the reasoning, identified policy conflicts, trade-offs and conditions, and facts gathered and used by the agent. An example of the resolution structure is shown below.

\lstdefinestyle{jsonbox}{
    basicstyle=\ttfamily\scriptsize,
    breaklines=true,
    breakatwhitespace=true,
    columns=fullflexible,
    keepspaces=true,
    showstringspaces=false,
    frame=none
}

\begin{tcolorbox}[
    colback=gray!8,
    colframe=gray!55,
    boxrule=0.5pt,
    arc=2mm,
    left=6pt,
    right=6pt,
    top=6pt,
    bottom=6pt,
    width=\linewidth
]
\begin{lstlisting}[style=jsonbox]
{
  "resolution_1": {
    "resolution_type": "RETURN_REFUND_FULL_BANK",
    "resolution_description": "Return approved only for the eligible SilkGlam Cleanse Makeup Remover Cloths. The mirror and foundation are excluded because they do not meet the original-condition policy. Refund will be issued to the original payment method after the returned item is received.",
    "customer_next_steps": "Print the prepaid label, attach it to the package, and drop it off within 30 days.",
    "agent_reasoning": "The customer requested to return three items, but only one item met the return-policy criteria."
  }
}
\end{lstlisting}
\end{tcolorbox}

\section{Human Annotation}
\label{app:human_annotation}
We collect human annotation at two stages of the project: at task creation (to confirm task realism and the possible set of outcomes). We use Label Studio~\citep{LabelStudio} to perform annotation. Each annotation and its process are described below.

\subsection{Task-Outcome Annotation}
\label{app:task_outcome_annotation}

To verify the quality of the tasks and their respective outcomes, we collect annotations for 10 (out of 40) representative tasks from three expert annotators. The annotations address two questions: (1) whether the task is realistic (i.e., plausible in the real world), and (2) what possible outcomes the task may have. The annotation followed the task generation, and stratification (Appendix~\ref{app:task_diversity}) is performed to confirm the quality of the tasks.

To annotate the tasks, we select them based on the diversity of 
ambiguities they embed and the diversity of products and customer reasons. Prior to annotation, annotators are provided with guidelines containing instructions for answering the annotation questions, along with the policy document they are expected to reference (Annotation guidelines given in Supplementary material document). During annotation, they are provided with task descriptions that include the product name, the customer's return reason, and the product condition. Additionally, each task specifies the situational context in which the return is initiated. An example task is shown below.

\begin{tcolorbox}[
    title={Example Task Description},
    colback=gray!6,
    colframe=gray!55,
    colbacktitle=gray!18,
    coltitle=black,
    fonttitle=\bfseries,
    boxrule=0.45pt,
    arc=1.5mm,
    left=6pt,
    right=6pt,
    top=6pt,
    bottom=6pt,
    width=\linewidth,
    enhanced,
    breakable
]
\small
Customer repeatedly purchases electronics and accessories and has recently returned several items, triggering a warning from Amazon about ``abuse of the return service.'' The latest order included: (1) ``SoundWaves Noise-Canceling Headphones,'' (2) ``ChargePro 30W USB-C Wall Adapter,'' (3) ``ProShield Anti-Glare Laptop Cover'' (all delivered 11 days ago). The headphones were opened and tested for about 2 hours, the wall adapter was used once, and the laptop cover remains unopened. Customer claims all items are in resalable condition with all packaging preserved, and states their only intent was to evaluate compatibility (no physical damage or wear). Customer is now requesting a return on all three items, having noticed a notice on their account about potential restrictions due to return frequency. The customer expresses concern and confusion over what ``reasonable use'' or ``abuse'' means, whether their current return is in jeopardy, and what threshold would trigger a return denial. They also ask if fees could be assessed due to ``obvious signs of use,'' raising questions about condition eligibility. The customer needs clarity on return rights and future account status since their need to test items before committing is ongoing and genuine.

\medskip
\noindent\textit{Do not assume facts not present in the task. Use conditional reasoning when information is missing.}
\label{task_description_example}
\end{tcolorbox}

The annotators reads the task and answers the following questions.
\begin{enumerate}
    \item Is this task realistic?
    \item Select all outcomes that a customer service agent might plausibly apply. Labels are checkboxes (multi-select). Then explain why the selected outcomes are plausible based on the task and policy.
\end{enumerate}

Every task had multiple plausible outcomes (mean 4.1, range 3--5), no task had a single agreed outcome, and the annotators disagreed on at least one outcome for 9 of 10 tasks (c.f. Table \ref{tab:per_task_agreement}). The mean exact annotation agreement for the tasks is 0.10, whereas the partial agreement, calculated using Jaccard similarity, is 0.46. We also observe that Escalate and Full Refund have the highest positive agreement (annotators often agree when they select them), which is expected, as these are the most ``default" outcomes.

\begin{table*}[t]
\centering

\small
\setlength{\tabcolsep}{5pt}
\renewcommand{\arraystretch}{1.15}

\begin{tabularx}{\textwidth}{@{}X l c c c@{}}
\toprule
\textbf{Task}
& \textbf{Agreement}
& \textbf{Human}
& \textbf{Cross-model}
& \textbf{Human--model} \\
&
\textbf{level}
& \textbf{(pairwise \%)}
& \textbf{(pairwise \%)}
& \textbf{(pairwise \%)} \\
\midrule

Multiple simultaneous home-goods orders (separate orders), with return
eligibility depending on how items from different orders are handled in
one return
& High
& 25\%
& 57\%
& 25\% \\

Order of three items from separate sellers in mixed conditions; return
eligibility depends on multi-seller bundles and partial use or condition
& Medium
& 25\%
& 33\%
& 27\% \\

Discounted three-pack of bundled T-shirts plus one pair of jeans,
delivered together, with one T-shirt problematic; return eligibility
hinges on the partial return of a discounted bundle
& Low 
& 12\%
& 10\%
& 16\% \\

\bottomrule
\end{tabularx}
\caption{Per-task agreement across human annotations, model outputs, and human--model comparisons.}
\label{tab:per_task_agreement}
\end{table*}

\subsection{Conversation Annotation}
\label{conv_annotation}

We use conversation annotation to calibrate the LLM-as-a-judge model (c.f. Section \ref{subsec:eval_pipeline}). To run this annotation, we select 10 tasks and collect the conversations between user and the agent LLMs. Then, the annotation task is formulated as follows: input: task, policy issues, conversation generated between the user agent and the customer service agent, containing dialogues and tool calls, and the resolution generated by the agent. Similar to task annotation, we provide a detailed annotation guideline with the policy documents, and we expect annotators to be well-versed in it (the guidelines are provided in the Supplementary Materials document). The annotators read the task, conversation, and resolution, and, based on these, answer the questions for the evaluation metrics below.

\begin{itemize}
    \item \textbf{CC1 · Conversation Consistency}: Are the agent's responses logically consistent with the customer's stated intent, previously shared information, and overall conversation context?

    \item \textbf{CC2 · Conversation Progression \& Resolution Mapping}: Does the conversation progress in a structured manner toward resolution, with each agent turn logically contributing to the final outcome? Consider both whether the conversation moves forward purposefully and whether the resolution is a direct, traceable result of what was discussed.

    \item \textbf{CC3 · Policy Adherence (Conversation)}: Does the agent's conduct across the full conversation align with the provided return policy and related policies? Consider whether the agent correctly applies the policy, omits relevant policy, or introduces unsupported / hallucinated policy clauses. Accuracy of what was said — not the final resolution; see OE3 for that.

    \item \textbf{CC4 · Verbosity}: Is the length and level of detail in the agent's responses appropriate for addressing the customer's needs? Repeated reassurance phrases with no new content, such as ``I understand your concern and I am here to help,'' across turns should score lower.

    \item \textbf{CC5 · Resolution Description Fidelity (Say-Do Gap)}: Does the resolution description accurately describe what was communicated to the customer, and does the conversation confirm the commitments reflected in the resolution?

    \item \textbf{CC6 · Interest Alignment: Customer}: To what degree does the agent's behavior and the outcome favor the customer's interest? Not a judgment of correctness. A score of 5 here does not imply a score of 1 on CC10.

    \item \textbf{CC7 · Interest Alignment: Company (Amazon)}: To what degree does the agent's behavior and the outcome protect the company's interest? Score independently of CC9.

    \item \textbf{OE1 · Resolution Type}: Select the single best label for the final action the agent took. Add a note below only if the refund method is unclear.

    \item \textbf{OE2a · Request Completeness}: Does the outcome address all aspects of the customer's stated return reason and request? Consider whether any part of what the customer asked for or described was overlooked, ignored, or left unresolved.

    \item \textbf{OE2b · Agent Effort / Going the Extra Mile}: Does the conversation and resolution reflect any effort by the agent to accommodate the customer beyond the minimum required by policy? Consider whether the agent proactively explored options, offered alternatives, or otherwise acted in the customer's interest without being prompted.

    \item \textbf{OE3 · Policy Adherence (Resolution)}: Is the final outcome clearly justified by the return policy given the task facts? Check: return window, item category, item condition, bundle rules, and non-returnable item list.

    \item \textbf{OE4 · Ambiguity Flagged}: Did the agent use discretion because the policy was unclear? Check whether the specific ambiguity is explicitly named in the Policy Issues list. Select ``Not applicable'' if the policy gave a clear answer.

    \item \textbf{OE5 · Reasoning Quality}: Does the outcome explanation demonstrate appropriate reasoning given the task complexity and policy issues? A more complex or ambiguous case should include more complete reasoning to score at the top of the scale.
\end{itemize}

As described earlier, we run these annotations with 3 expert annotators, each of whom assigned 15 tasks, yielding 15 tasks with 2 annotations. The annotation agreement between the annotator pairs is shown in Table~\ref{tab:annotator_agreement}. Similar to \citet{kawarada2026gain}, we consider this disagreement as a result of the complexity of the benchmark, as it reflects the lack of a single correct answer across dimensions. It is also consistent with annotators converging on the broad direction of a judgment while differing on the specific integer point on the 5-point scale, rather than disagreeing on the underlying agent quality. We acknowledge that small sample size and inherent Likert-rubric subjectivity also contribute to the low exact agreement. 

\begin{table}[t]
\centering
\footnotesize
\setlength{\tabcolsep}{3pt}
\renewcommand{\arraystretch}{1.08}

\begin{tabularx}{\columnwidth}{@{}Xrrrr@{}}
\toprule
\textbf{Annotators } & \textbf{Tasks} & \textbf{Exact IAA} & \textbf{Macro $\kappa_w$} & \textbf{Macro $\alpha_K$} \\
\midrule
Ann1 vs Ann2 & 5  & 0.33 & $-0.04$ & $-0.16$ \\
Ann1 vs Ann3 & 5  & 0.27 & $+0.02$ & $-0.15$ \\
Ann2 vs Ann3 & 5  & 0.09 & $+0.04$ & $-0.06$ \\
\midrule
\textbf{Overall} & \textbf{15} & \textbf{0.23} & $\mathbf{+0.04}$ & $\mathbf{-0.086}$ \\
\bottomrule
\end{tabularx}

\vspace{0.4em}
\caption{Inter-annotator agreement across annotator pairs. Exact IAA denotes mean exact agreement; Macro $\kappa_w$ and Macro $\alpha_K$ are means of per-metric values across the 11 ordinal metrics. IAA is computed on raw annotator labels prior to adjudication.}

\label{tab:annotator_agreement}
\end{table}

\section{LLM-as-Judge Calibration Results}
\label{app:judge_calibration}
In this section, we discuss the error metrics and agreements by evaluation dimension that we use for LLM-as-Judge calibration. The calibration matrices that we use are defined below.

\begin{itemize}
    \item Agreement (exact agreement): Fraction of scenarios where the judge picked the exact same integer as the human. The strictest match — no partial credit. 

    \item Within-1 (adjacent agreement): Fraction where the judge is at most one ordinal step off (e.g., human=4, LLM=3 or 5). Allows for the unavoidable rubric ambiguity between adjacent levels of a 5-point scale.

    \item MAE (mean absolute error): Average magnitude of disagreement, in score points. Direction-agnostic. Penalizes a 2-point miss twice as much as a 1-point miss.

    \item Bias: Signed mean difference. Tells the direction of error: positive = LLM is systematically lenient (scores higher than humans); negative = LLM is strict (scores lower).
    
\end{itemize}

Table \ref{tab:agreement_metrics_by_dimension} contain the results of the agreement and error metrics. The mean exact annotation agreement post calibration is \textbf{0.45}, which although is moderate, can be considered acceptable due to subjectivity of the metrics.

The calibration ground-truth label is the adjudicator consensus where two annotators were assigned (15 of 29 scenarios); for the remaining 14 singly-annotated scenarios, the single annotator's label is used. Where annotators disagreed and adjudication did not explicitly resolve the disagreement ($\approx 75\%$ of disagreement rows), the higher of the two annotator labels is taken — a deliberately conservative choice that biases the ground truth toward the more generous reading. Inter-annotator agreement in Table~\ref{tab:annotator_agreement} measures \emph{raw} pairwise human-vs-human concordance prior to any adjudication and reflects the inherent subjectivity of multi-dimensional Likert rubrics on policy-conflict scenarios. The LLM judge is calibrated against the post-adjudication consensus, not against any single annotator's labels.

\begin{table}[t]
\centering
\scriptsize
\setlength{\tabcolsep}{2.5pt}
\renewcommand{\arraystretch}{1.08}

\begin{tabularx}{\columnwidth}{@{}p{0.22\columnwidth}Xrrrr@{}}
\toprule
\textbf{Dimension} & \textbf{Metric} & \textbf{Agr.} & \textbf{W-1} & \textbf{MAE} & \textbf{Bias} \\
\midrule

\multirow{3}{=}{Policy Adherence}
& Policy Support (Conversation) & 0.36 & 0.88 & 0.76 & $+0.04$ \\
& Policy Support (Resolution) & 0.44 & 0.80 & 0.80 & $+0.08$ \\
& Reasoning Quality & 0.42 & 0.83 & 0.63 & $-0.29$ \\
\midrule

\multirow{4}{=}{Dialogue Quality}
& Verbosity & 0.68 & 0.84 & 0.52 & $+0.44$ \\
& Resolution-Oriented Responses & 0.44 & 0.96 & 0.60 & $+0.12$ \\
& Resolution Description Fidelity & 0.52 & 0.72 & 0.88 & $+0.88$ \\
& Conversation Consistency & 0.32 & 0.76 & 0.92 & $+0.84$ \\
\midrule

Task Resolution
& Addresses Customer Reason & 0.71 & 0.92 & 0.38 & $+0.38$ \\
\midrule

\multirow{2}{=}{Interest Alignment}
& Customer Interest Alignment & 0.28 & 0.80 & 0.92 & $+0.12$ \\
& Company Interest Alignment & 0.36 & 0.84 & 0.80 & $+0.16$ \\
\midrule

\textbf{Mean} &  & \textbf{0.45} & \textbf{0.84} & \textbf{0.72} & $\mathbf{+0.33}$ \\
\bottomrule
\end{tabularx}

\vspace{0.4em}
\caption{Agreement and error metrics by evaluation dimension and metric. Agr. denotes exact agreement; W-1 denotes within-one agreement.}
\label{tab:agreement_metrics_by_dimension}
\end{table}

\section{Experiment Setting}
\label{app:expt_setting}
 In this section, we describe the experimental setting of our agentic setup. We maintain the same setup across the models. Table \ref{tab:expt_settings}.

\begin{table}[h]
\centering
\small
\setlength{\tabcolsep}{6pt}
\renewcommand{\arraystretch}{1.2}
\begin{tabular}{@{}lr@{}}
\toprule
\textbf{Parameter} & \textbf{Value} \\
\midrule
\multicolumn{2}{@{}l}{\textit{Shared across all models}} \\
\midrule
Customer model         & \texttt{gpt-4o-mini} \\
Agent temperature      & 0.7 \\
Customer temperature   & 0.8 \\
Max turns              & 20 \\
Max output tokens/turn & 4{,}096 \\
Tool calling mode      & Native / JSON-prompt$^{\dagger}$ \\
\midrule
\multicolumn{2}{@{}l}{\textit{Local models (Qwen3.6-35B-A3B / gpt-oss-120b)}} \\
\midrule
CPU cores              & 8 / 16 \\
GPU count              & 1$\times$ / 4$\times$ \\
GPU type               & H100 \\
Run time limit         & 96 h \\
\bottomrule
\end{tabular}
\caption{Model execution parameters. Top block: shared across all models. Bottom block: applies to local models only; API models use provider defaults. $^{\dagger}$Native tool calling for API models, JSON-prompt for local models.}
\label{tab:expt_settings}
\end{table}

 We use the LiteLLM \footnote{\url{https://github.com/BerriAI/litellm}} package to structure and collect the conversation for the API models, and we use vLLM \citep{vllm2023} for the inference backend for the local models.  As for the models, we add the model information in Table \ref{tab:model_info} containing name, size, provider, and license information.

\begin{table}[t]
\centering
\footnotesize
\setlength{\tabcolsep}{4pt}
\begin{tabular}{@{}llll@{}}
\toprule
\textbf{Model} & \textbf{API / Model ID} & \textbf{Size} & \textbf{License} \\
\midrule
GPT-4o mini    & gpt-4o-mini       & --                & Prop. \\
GPT-4.1        & gpt-4.1           & --                & Prop. \\
GPT-5          & gpt-5             & --                & Prop. \\
GPT-5.4        & gpt-5.4           & --                & Prop. \\
GPT-OSS 120B   & gpt-oss-120b      & 117B/5.1B$^{*}$   & Apache \\
Qwen 3.6 35B   & Qwen3.6-35B-A3B   & 36B/3B$^{*}$      & Apache \\
\bottomrule
\end{tabular}
\caption{Overview of selected language models. All OpenAI models accessed via API; open-source models (GPT-OSS, Qwen) hosted via Hugging Face. $^{*}$Total/active parameters (MoE). Prop. = Proprietary.}
\label{tab:model_info}
\end{table}

\section{Customer Agent Simulation Setup Description}
\label{app:conv_simulation}
We will now describe the setup of the simulation that we adapt in designing the conversation simulation between the customer and the agent. We follow the standard dual-control conversation framework from \citet{barres2025tau}, with certain modifications. The process is described below.

\begin{itemize}
    \item Step 0: Task Split: At scenario generation, each task is split into two asymmetric briefs: 1) Agent task detail: A CRM style framing of the task., 2) Customer task detail: Containing information that customer will share, behavioral information, facts that customer holds back until asked, and the customer first message.
    
    \item Step 1: Load scenario and build components (orchestrator): er variant, instantiate Agent, Customer, and Environment from the same scenario.

    \item Step 2: Seed the conversation: The pre-generated customer message is appended to first customer turn.

    \item Step 3: Agent Turn : Each agent turn the LLM sees: agent persona (DIRECT/FAIR/AGREEABLE/HELPFUL/VERY HELPFUL), primary and related policies, details, full conversation history (customer's tool results hidden). The agent can: 1) Emit a customer-facing message (clarifying questions), 2) Call read tools, 3) Call write tools.

    \item Step 4: Customer Turn: Now it is the simulated customer's turn to respond. The customer model is given a complete picture of the situation, the entire conversation and the agent's most recent message. 

    \item  Step 5: Loop until terminal:  Steps 4 and 5 (the agent's turn followed by the customer's turn)keep repeating, building the conversation back and forth. The loop only stops when one of four things happens: the agent reaches a final resolution and successfully closes the case, the customer decides to walk away from the conversation, or the conversation hits the maximum allowed number of turns (ten by default). Whichever condition trips first, the variant is considered done.

    \item Step 6: Flatten and Save: Once the conversation reaches its end, it is flattened into a single output record and saved to a JSONL file with one line per scenario, in exactly the format the downstream evaluation pipeline expects to consume.

\end{itemize}

\begin{figure*}[b]
    \centering
    \includegraphics[width=\textwidth]{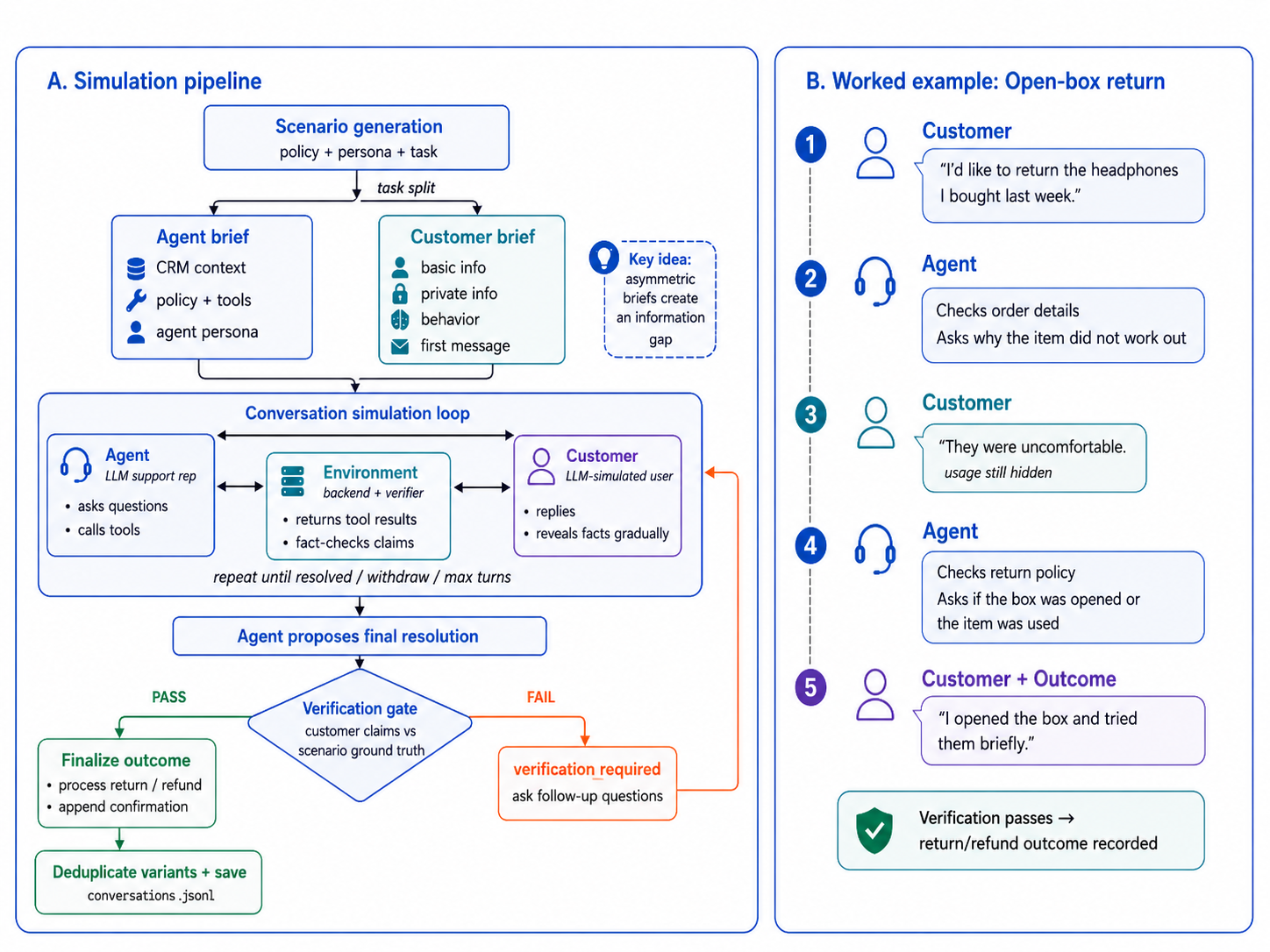}
    \caption{
    Overview of the evaluation pipeline. Each of the judges (implemented in parallel) represents the evaluation dimensions.
    }
    \label{fig:simulation_pipeline}
\end{figure*}

The entire pipeline is illustrated in Figure \ref{fig:simulation_pipeline}, along with an example to show how it works. The figure is quite high-level and omits technical details for better understanding.

\section{Cross-Domain Generalization with Airline Domain}
\label{app:cross_domain_airline}

To evaluate generalizability, we ported DRIP-R to the $\tau2$ airline domain \cite{barres2025tau}. Our pipeline (policy ambiguity extraction, scenario generation, agent simulation, and judge-free outcome measurement) remained unchanged; only domain-specific information, including the policy, tools, prompts, and outcome taxonomy, was replaced.

Because $\tau2$ tasks are designed to admit a single gold outcome under $\tau2$'s precisely specified policy, we paired 50 $\tau2$ airline tasks with a consolidated policy derived from Ryanair's real-world policies. We then used GPT-4.1 to verify that each task admitted multiple plausible outcomes under this new policy. To assess annotation quality, we randomly sampled five tasks and manually reviewed their outcome annotations, which were generally satisfactory. Each task was paired with two personas, yielding 100 scenarios in total.

Given time constraints, we evaluated two models, GPT-4.1 and GPT-5, with three runs per model on each scenario. We obtained the following main findings. First, the judge-free metric transferred successfully to the new domain and distinguished between the models: GPT-4.1 was nearly deterministic, varying across identical reruns in only 16 of 100 scenarios, whereas GPT-5 varied in 46 of 100 scenarios, or 23 of 100 after near-synonymous outcome labels were collapsed. Second, the two models selected different resolutions in 47\% of scenarios. We consider this figure to be a conservative estimate, as the original τ² airline tasks are generally much simpler than the retail-domain tasks introduced in our paper, where greater task complexity may create more scope for divergent resolutions.






\end{document}